\documentclass{article} %
\usepackage{iclr2027_conference,times}

\usepackage{enumitem}
\usepackage{multirow}
\usepackage{wrapfig}
\usepackage{colortbl}
\usepackage[normalem]{ulem}
\usepackage{microtype}
\usepackage{comment}
\usepackage{pifont}
\usepackage{xspace}
\newcommand{\cmark}{\ding{51}}
\newcommand{\xmark}{\ding{55}}

\setlist[itemize]{align=parleft,left=0pt,topsep=1mm,itemsep=0mm,parsep=1mm}

\definecolor{azure(colorwheel)}{rgb}{0.0, 0.5, 1.0}
\definecolor{nicegreen}{rgb}{0.0, 0.7, 0.1}
\definecolor{yw}{rgb}{0.01176, 0.5490, 0.5490}
\definecolor{ashblue}{rgb}{0.36, 0.54, 0.66}
\definecolor{ashgrey}{rgb}{0.7, 0.75, 0.71}
\definecolor{applegreen}{rgb}{0.55, 0.71, 0.0}
\definecolor{blue}{rgb}{0.0, 0.0, 1.0}
\definecolor{postechred}{rgb}{0.784, 0.003, 0.313}
\definecolor{ywg}{rgb}{0.9960, 0.8984, 0.5859}
\definecolor{ballblue}{rgb}{0.13, 0.67, 0.8}
\definecolor{cornellred}{rgb}{0.7, 0.11, 0.11}
\definecolor{darkcyan}{rgb}{0.0, 0.55, 0.55}
\definecolor{CuGray}{gray}{0.9}
\definecolor{airforceblue}{rgb}{0.36, 0.54, 0.66}
\definecolor{rev}{rgb}{0.784, 0.003, 0.313}
\definecolor{pink}{cmyk}{0, 0.7808, 0.4429, 0.1412}
\definecolor{amethyst}{rgb}{0.6, 0.4, 0.8}
\definecolor{black}{rgb}{0.0, 0.0, 0.0}
\definecolor{tb3_yellow}{rgb}{0.996, 1.0, 0.6}
\definecolor{tb3_orange}{rgb}{0.980, 0.8, 0.604}
\definecolor{tb3_red}{rgb}{0.972, 0.6, 0.6}
\definecolor{dimgray}{rgb}{0.41, 0.41, 0.41}
\definecolor{brickred}{rgb}{0.8, 0.25, 0.33}
\definecolor{bleudefrance}{rgb}{0.19, 0.55, 0.91}
\definecolor{blue(ncs)}{rgb}{0.265, 0.445, 0.765}
\definecolor{blue(ryb)}{rgb}{0.01, 0.28, 1.0}
\definecolor{orange}{rgb}{1.0, 0.49, 0.0}
\definecolor{Gray}{gray}{0.88}
\definecolor{green(ncs)}{rgb}{0.0, 0.62, 0.42}
\definecolor{brightpink}{rgb}{1.0, 0.0, 0.5}
\definecolor{pastelred}{rgb}{0.66, 0.25, 0.28}
\definecolor{pastelorange}{rgb}{0.54, 0.38, 0.30}
\definecolor{pastelgreen}{rgb}{0.39, 0.55, 0.38}
\definecolor{pastelblue}{rgb}{0.34, 0.42, 0.90}
\definecolor{pastelpurple}{rgb}{0.50, 0.30, 0.50}
\definecolor{alizarin}{rgb}{0.82, 0.1, 0.26}
\definecolor{darkred}{rgb}{0.7,0.2,0.1}
\definecolor{nvgreen}{rgb}{0.573, 0.776, 0.200}

\definecolor{c_diff}{rgb}{0.3058823529, 0.5843137255, 0.8509803922}
\definecolor{c_urn}{rgb}{0.98, 0.4941176471, 0.4745098039}
\definecolor{c_upm}{rgb}{0.3058823529, 0.6549019608, 0.1803921569}
\definecolor{c_exp}{rgb}{1.0, 0.7529411765, 0.0}
\definecolor{iccvblue}{rgb}{0.21,0.49,0.74}
\definecolor{pink}{cmyk}{0, 0.7808, 0.4429, 0.1412}

\definecolor{kellygreen}{rgb}{0.3, 0.73, 0.09}
\newcommand{\colorref}[1]{{\color{cornellred}{#1}}}

\newcolumntype{g}{>{\columncolor{CuGray}}c}
\newcolumntype{z}{>{\columncolor{CuGray}}l}

\renewcommand{\paragraph}[1]{\vspace{1mm}\noindent\textbf{#1.}\,\,}

\newcommand{\greencap}[1]{\textcolor{kellygreen}{#1}}

\newcommand{\red}[1]{\textcolor{alizarin}{#1}}

\definecolor{tabfirst}{rgb}{1, 0.7, 0.7} %
\definecolor{tabsecond}{rgb}{1, 0.85, 0.7} %
\definecolor{tabthird}{rgb}{1, 1, 0.7} %

\def\ours{NeuMatEx\xspace}

\usepackage{xspace}

\makeatletter
\def\@fnsymbol#1{\ensuremath{\ifcase#1\or *\or \dagger\or \ddagger\or
   \mathsection\or \mathparagraph\or \|\or **\or \dagger\dagger
   \or \ddagger\ddagger \else\@ctrerr\fi}}
\makeatother

\def\onedot{.\@\xspace}
\def\eg{\emph{e.g}\onedot} 
\def\ie{\emph{i.e}\onedot}

\newcommand{\Sref}[1]{\cref{#1}}
\newcommand{\Eref}[1]{Eq.~\ref{#1}}
\newcommand{\Fref}[1]{\cref{#1}}
\newcommand{\Tref}[1]{\cref{#1}}

\newcommand{\bp}{{\mathbf{p}}}

\newcommand{\bz}{{\mathbf{z}}}

\newcommand{\bI}{\mathbf{I}}

\newcommand{\bT}{\mathbf{T}}

\newcommand{\calT}{{\mathcal{T}}}

\newcommand{\be}{\begin{eqnarray}}
\newcommand{\ee}{\end{eqnarray}}
\newcommand{\bee}{\begin{eqnarray*}}
\newcommand{\eee}{\end{eqnarray*}}

\newcommand{\matrixb}{\left[ \begin{array}}
\newcommand{\matrixe}{\end{array} \right]}

\newcommand{\argmin}{\operatornamewithlimits{\arg \min}}

\newcommand{\E}{\mathbb E}

\newcommand{\videoInput}{\mathbf{I}}

\newcommand{\trilat}{\mathbf{tri}}
\newcommand{\diffusionModel}{\mathcal{F}}
\newcommand{\diffusionModelParams}{\theta}
\newcommand{\diffusionModelFn}{\diffusionModel_{\diffusionModelParams}}
\newcommand{\vaeEncoder}{\mathcal{E}_\text{VAE}}
\newcommand{\vaeDecoder}{\mathcal{D}_\text{VAE}}

\newcommand{\mlpmat}{\mathcal{M}_{\phi}^{\text{mat}}}
\newcommand{\mlpunc}{\mathcal{M}_{\psi}^{\text{unc}}}

\newcommand{\dataDistribution}{p_{\text{data}}}

\newcommand{\std}[1]{{\scriptsize$\pm$#1}}
\newcommand{\nmLatent}{\ell}
\newcommand{\nmDecoder}{\mathcal{D}_\text{neu}}
\newcommand{\gbuffermat}{\mathbf{G}_{\text{mat}}}
\newcommand{\gbufferunc}{\mathbf{G}_{\text{unc}}}
\newcommand{\Txy}{\mathbf{T}_\mathrm{XY}}
\newcommand{\Tyz}{\mathbf{T}_\mathrm{YZ}}
\newcommand{\Txz}{\mathbf{T}_\mathrm{XZ}}
\newcommand{\redcross}{\red{\xmark}}
\newcommand{\greencheck}{\greencap{\cmark}}

\definecolor{cvprblue}{rgb}{0.37,0.59,0.79}
\usepackage[pagebackref,breaklinks,colorlinks,citecolor=cvprblue,linkcolor=cornellred]{hyperref}
\usepackage{url}
\usepackage{graphicx}
\usepackage{caption}
\usepackage{booktabs}
\usepackage{amsmath}
\usepackage{amssymb}
\usepackage[capitalize]{cleveref}
\usepackage{array}
\usepackage{tabularx}
\usepackage{adjustbox}
\usepackage[hang,flushmargin]{footmisc}
\usepackage{xcolor}
\usepackage{subcaption}

\crefname{section}{Sec.}{Secs.}
\Crefname{section}{Section}{Sections}
\Crefname{table}{Table}{Tables}
\crefname{table}{Tab.}{Tabs.}

\title{Extracting Neural Materials from Images}

\author{%
Kim Youwang$^{1,2}$ \quad Jon Hasselgren$^{1}$ \quad Peter Kocsis$^{1}$ \\
\textbf{Andrea Weidlich}$^{1}$ \quad \textbf{Tae-Hyun Oh}$^{3}$ \quad \textbf{Jacob Munkberg}$^{1}$
\\
\newline
\\
$^{1}$NVIDIA \quad $^{2}$POSTECH \quad $^{3}$KAIST
\\
\,\,\url{https://nvlabs.github.io/neumatex/}\vspace{-3mm}
}

\iclrfinalcopy %
\begin{document}

\makeatletter
  \let\savedmaketitle\maketitle \let\savedmk\@maketitle \let\savedauthor\@author
  \makeatother
{\maketitle}

\vspace{-4.5mm}
\begin{figure*}[h]
\centering
\includegraphics[width=\linewidth]{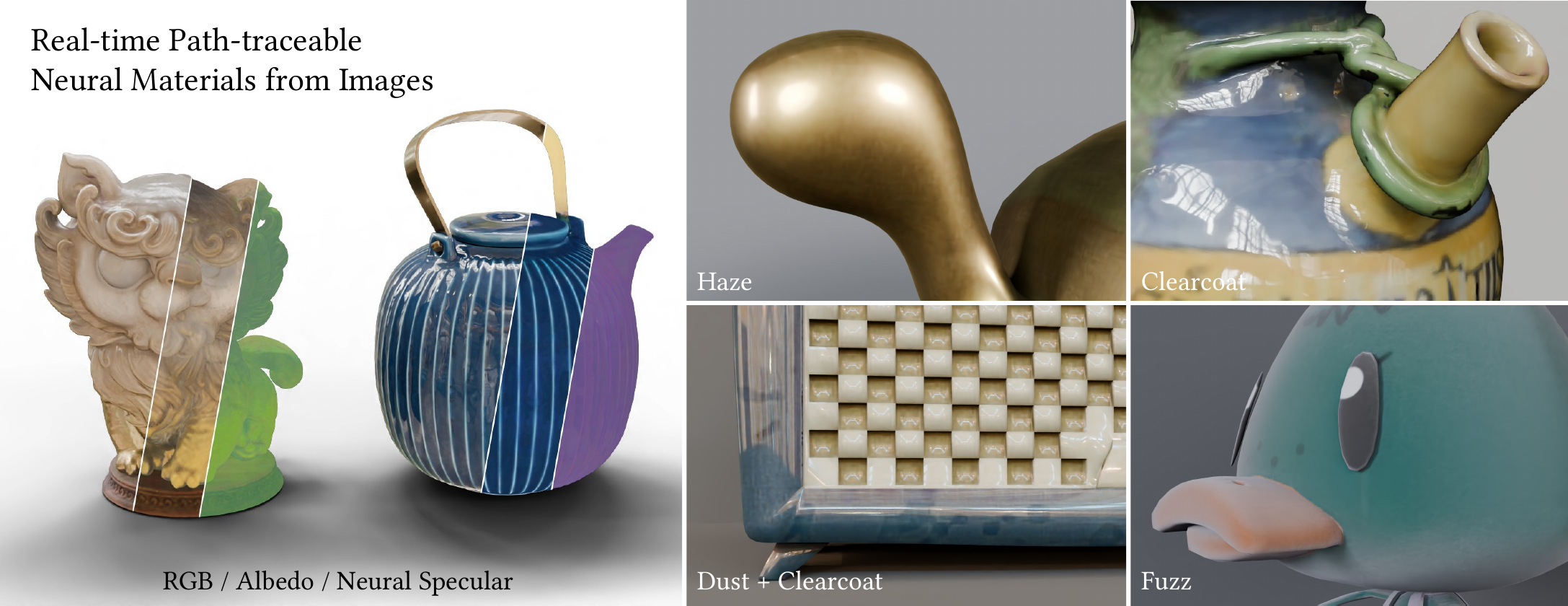}\vspace{-2mm}
\caption{\textbf{\ours} extracts \emph{Neural Materials}, a richer material representation that goes beyond PBR, from multi-view images. 
\ours decomposes a diffuse lobe and neural specular lobes to model complex material-light interactions such as \emph{haze}, \emph{dust}, \emph{clearcoat}, \emph{fuzz}, \emph{scatter}, and their mixtures. The extracted neural materials support real-time path tracing on a commodity GPU.
}
\label{fig:teaser}
\end{figure*}

\begin{abstract}
Neural materials can represent complex specular reflections and scattering 
effects in a compact, universal basis. However, acquiring and authoring such 
materials remains challenging. 
We present \ours, a differentiable inverse rendering method for extracting 
spatially varying neural materials from images. The nonlinear structure of 
neural material latent spaces makes optimization with na\"ive inverse 
rendering infeasible. 
To address this, we train a Large Material Reconstruction Model~(LMRM) that 
directly predicts initial base color, neural material latents, and aleatoric 
uncertainty guides from images. This material prior provides a good initialization 
and better constrains our subsequent optimization using inverse path tracing. 
The predicted uncertainty further helps by anchoring high-confidence regions 
more tightly to the LMRM prediction, preventing lighting and complex specular 
effects from being baked into materials. 
Experiments on synthetic and real assets show that \ours extracts complex materials 
with better visual quality and material decomposition than PBR-based methods.
\end{abstract}

\addtocontents{toc}{\protect\setcounter{tocdepth}{-1}}
\section{Introduction}
\label{sec:intro}

Authoring materials for 3D objects is a challenging artistic task, which involves creating 
spatially-varying details and assigning appropriate material parameters to their different parts.
Current interactive applications, \eg, games, commonly apply a physically-based (PBR) material model, 
where a material is defined by diffuse base color, roughness, and metallicity stored in spatially varying textures. 
This allows a reasonable variety of materials, 
including metals, woods, and plastics, but is still far from the appearance quality in offline rendering,
\eg, film production, where artists design large material node graphs, \eg, MaterialX~\citep{materialx}, with multiple layers to capture the 
real-world appearance of effects like skin, eyes, and car paint.

To bridge the gap between offline and real-time material quality, \emph{neural} material 
representations~\citep{kuznetsov2021neumip,fan2022nlb,zeltner2024neural,yu2026toward} have recently been proposed. 
The idea is to bake complex material graphs into a compact neural representation, consisting of a set 
of latent texture maps and a small neural network.
This results in a unified format efficiently executable on modern graphics hardware at real-time rates, 
\eg, inside a path tracing real-time renderer~\citep{zeltner2024neural}.
Neural materials are typically baked from large material node graphs by densely sampling the 6D space 
of spatial variation (2D) and incoming and outgoing directions (4D). 
For each sample, the spatially varying bidirectional reflection/scattering function (SVBSDF) is evaluated 
and the neural material is optimized to match the reference material's SVBSDF as closely as possible.

In this work, we ask whether the inverse rendering machinery can be extended to extract \emph{neural material} 
representations, which offer far greater expressiveness.
To help asset creation, we introduce the first pipeline to extract neural materials from multi-view images. 
Recent works have successfully extracted PBR materials from multi-view images via inverse rendering, leveraging the highly constrained PBR representation.
Neural materials, on the other hand, are much more expressive, with additional degrees of freedom, and 
the latent space representing the manifold of valid materials can be fractured.
This makes inverse rendering a highly non-convex problem with many poor local minima, where the 
optimization can succeed only when initialized within the basin of attraction of a good solution.
We address this problem by combining feed-forward reconstruction model priors with differentiable 
optimization to robustly extract neural materials from images.

Specifically, we have constructed a large feed-forward material reconstruction model (LMRM) which 
produces 1) a high-quality initial guess of the neural material from a set of image observations, 
and 2) uncertainty estimates of the predictions. 
In a second step, we further refine the material prediction using Monte Carlo-based inverse path tracing, guided by uncertainty. 
We leverage a recent neural material basis~\citep{yu2026toward} which focuses on high-quality specular appearance, including clearcoat, dust, fuzz, and scatter effects (see \Fref{fig:teaser}). 
Our extracted neural materials can be directly deployed inside a real-time path tracer.
We summarize our main contributions as follows:

\begin{itemize}
\item We introduce the first image-based neural material extraction for 3D objects, combining a Large Material Reconstruction Model (LMRM) with Test-Time Optimization (TTO), outperforming PBR.
\item We train an LMRM to estimate neural materials from multi-view images, providing good initialization for our subsequent TTO with inverse path tracing.
\item Our uncertainty-guided material regularization better constrains TTO and prevents baked-in effects. 
\end{itemize}

\section{Related Work}
\label{sec:related}

Our work targets extracting neural materials from images by combining inverse rendering with a large feed-forward reconstruction model. We briefly review the related fields.

\paragraph{\textbf{Material extraction from images}}
A long-standing goal in inverse rendering is to extract spatially varying materials from 
images of 3D shapes. Several techniques estimate surface radiometric properties from images. 
Previous works on SVBRDF estimation rely on special viewing configurations, lighting patterns or complex capturing setups~\citep{hendrik2003planned,Gardner2003,Ghosh2009,Guarnera2016,Weinmann2015,bi2020neural,boss2020two,Schmitt2020CVPR}. 
More recently, optimization-based methods jointly recover PBR texture maps, environment lighting and
geometry from casually captured images~\citep{azinovic2019ipt,StyleGAN3D,chen2021dibrpp,munkberg2022nvdiffrec,boss2021neural,hasselgren2022nvdiffrecmc,zhang2022modeling,lin2025iris} using inverse rendering.
Radiance field-based methods factorize scene representations into intrinsic components without relying 
on explicit geometry~\citep{zhang2021nerfactor,jin2023tensoir,jiang2024gaussianshader,gao2024relightable, liang2024gs}. 

Optimization-based methods are sensitive to ambiguous data and often require well-calibrated  
observations to reliably disentangle lighting, reflectance, and shape.
To overcome this, a plethora of methods rely on neural networks to predict BRDFs from images~\citep{Gao2019,Guo2020,li2020inverse,li2018learning,nimier2021,Luan2021}.
Additionally, many recent methods use feed-forward or generative approaches to directly predict material intrinsics~\citep{kocsis2024iid,zeng2024rgbx,litman2025materialfusion,youwang2024paintit,munkberg2025videomat,hasselgren2026videomatgen,kocsis2026iif,li2025lirm,siddiqui2024assetgen,hunyuan3d21_2025_hunyuan3d,xiang2026trellis2,li2026lsrm}. 
While the field has advanced substantially, these methods universally target standard PBR material representations, \ie, 
a diffuse term paired with a single-lobe specular component, which limits the complexity of the extracted materials.

\paragraph{\textbf{Neural materials}}
Neural material models represent SVBSDFs using latent textures for spatial variation
and neural decoders for directional variation~\citep{kuznetsov2021neumip,fan2022nlb,zeltner2024neural}.
To improve rendering performance, the decoders are highly optimized MLPs designed around hardware 
constraints~\citep{zeltner2024neural}, but the complexity of the compressed latent space typically 
requires special consideration during optimization~\citep{bitterli2026taming}.

Recent work explores neural-material generation through diffusion-based synthesis of
neural textures~\citep{raghavanmullia2025genneumat} or reconstruction from generative
video priors~\citep{xue2026videoneumat}. 
VideoNeuMat~\citep{xue2026videoneumat} focuses on obtaining planar 2D neural materials from generated videos,
rather than recovering materials on 3D objects from images or real-world photos. 
In contrast, we combine object-level inverse rendering with the richer neural material basis of~\citet{yu2026toward}. 
Since the decoder maps latent textures to appearance nonlinearly, test-time optimization is
sensitive to initialization. We therefore learn a large-scale image-to-neural-material reconstruction model for initialization and
refine its prediction using differentiable path tracing.

\paragraph{\textbf{Relightable neural representations}} 
Several works use neural representations for relighting surface and volumetric 
assets~\citep{mullia2024rna, zeng2023nrhints, gilles2022nprt, fan2025rng, yingyan2023renerf}. 
They typically treat assets as atomic units and precompute light transport, which makes animation, 
local lighting, and object interaction difficult. 
For instance, reconstructed objects generally do not support mutual global illumination effects. 
In addition, asset generation often requires controlled-lighting setups~\citep{mullia2024rna, fan2025rng, yingyan2023renerf}, 
typically with point light sources, complicating real-world data capture. 
In contrast, our method relies on inverse rendering~\citep{munkberg2022nvdiffrec,boss2021neural,hasselgren2022nvdiffrecmc,zhang2022modeling} 
and explicitly reconstructs the local SVBSDF, making the resulting assets compatible with off-the-shelf 
animation and light-transport algorithms.

\section{Overview of \ours}
\label{sec:overview}

\paragraph{Neural Materials (NM)}
Unlike standard PBR materials~\citep{burley2012physically,karis2013real} parameterized by 
explicit maps (base color, roughness, metallicity),
a Neural Material (NM)~\citep{kuznetsov2021neumip,fan2022nlb,zeltner2024neural}
represents a richer SVBSDF with a set of latent textures and a small neural network.
We adopt the recent neural material SVBSDF of \citet{yu2026toward}, which combines a Lambertian diffuse lobe 
with an expressive \emph{neural} specular lobe, evaluated at a surface point $\bp$ with incoming and outgoing directions $\omega_i$, $\omega_o$:
\begin{equation}
    \label{eq:svbsdf}
    f(\bp, \omega_i, \omega_o) = T_\mathrm{neu}(\bp, \omega_i)\,\frac{\rho_d(\bp)}{\pi} + f_\mathrm{neu}(\bp, \omega_i, \omega_o),
\end{equation}
where $\rho_d$ is the diffuse base color, $f_\mathrm{neu}$ the specular BSDF, and $T_\mathrm{neu}$ 
the transmission albedo enforcing energy conservation.
The latter two are
decoded by a learned universal decoder MLP $\nmDecoder$:
\begin{equation}
    \label{eq:nm_fwd}
    (f_\mathrm{neu}, T_\mathrm{neu}, R_\mathrm{neu}) = \nmDecoder(\nmLatent(\bp),\, \omega_i, \omega_o),
\end{equation}
where $\nmLatent(\bp)\in\mathbb{R}^{6}$ is a per-point neural specular latent code that compresses 22 analytical 
SVBSDF parameters~\citep{yu2026toward}, 
and $R_{\mathrm{neu}}$ is a scalar measure of reflected energy of $f_{\mathrm{neu}}$, used to importance-sample between the diffuse and neural specular lobes during rendering.
Learned from a large-scale 3D shape dataset with procedurally enhanced NM textures~\citep{yu2026toward} spanning complex multi-lobe specular effects, \eg, clearcoat and scattering, this NM latent space covers a far richer appearance than PBR's single GGX lobe.

In this work, we aim to extract $\rho_d(\bp)$ and $\nmLatent(\bp)$ from images. Since $\nmLatent(\bp)$ is a neural latent observed through the nonlinear decoder $\nmDecoder$, not a physical parameter, extracting it is highly non-convex; we thus predict both with a large feed-forward model and refine them via test-time optimization.

\begin{figure*}[t]
\centering
\includegraphics[width=\linewidth]{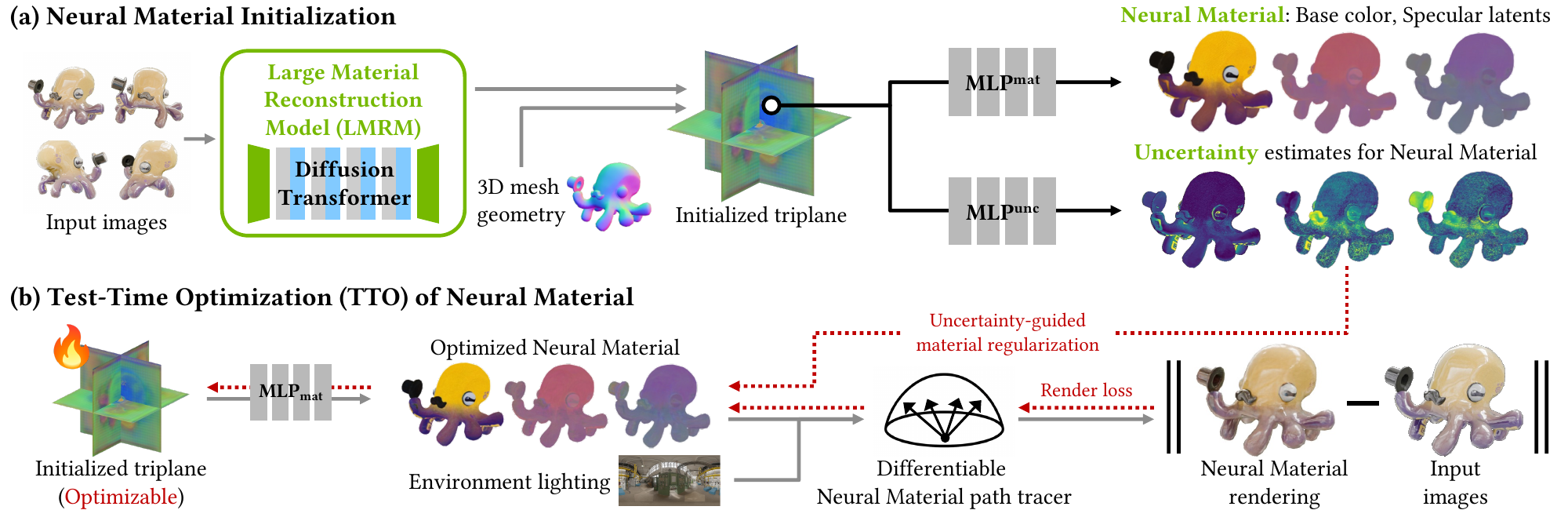}
\caption{\ours consists of two stages.
\textbf{(a) Neural Material Initialization:} Given input images,
a Large Material Reconstruction Model (LMRM) predicts a feature triplane in a single forward pass. The triplane is decoded by two MLPs to jointly predict an initial neural material and per-material uncertainty that reflects ambiguous surface regions.
\textbf{(b) Test-Time Optimization (TTO):} We further optimize the initialized triplane via differentiable path tracing supervised by a photometric render loss against the input images. To regularize the ill-posed inverse rendering, uncertainty estimates from (a) are used to guide the TTO, allowing flexible material drifts for uncertain regions and preventing significant material drifts for confident regions, steering the solution away from local minima.
}
\label{fig:system}
\end{figure*}

\paragraph{\ours Pipeline}
\ours consists of two stages (see \Fref{fig:system}).
In the first stage (\Fref{fig:system}\colorref{(a)}), a Large Material Reconstruction Model (LMRM) predicts 
a feature triplane from multi-view images and mesh geometry, which is decoded into an initial neural 
material and per-point uncertainty estimates.
In the second stage (\Fref{fig:system}\colorref{(b)}), we refine the triplane via differentiable inverse path tracing, using the predicted uncertainty to regularize the ill-posed material decomposition.

\section{Large Material Reconstruction Model}
\label{sec:ff_init}
\label{sec:lmrm}

We introduce a large feed-forward material reconstruction model (LMRM) that takes multi-view images and 
jointly predicts 1) a high-quality initial guess of the neural material, and 2) uncertainty estimates of 
the material predictions.

\paragraph{Model architecture}
Following recent work~\citep{wei2024meshlrm,zhang2024relitlrm,xue2026videoneumat}, we leverage a pre-trained text-to-video
diffusion transformer (DiT)~\citep{peebles2022dit}, Wan2.1-1.3B~\citep{wan2025wan}, repurposed 
as a \emph{single-step} model. 
We fine-tune the model to take a set of input views, $\videoInput$, and to output \emph{neural materials} 
parameterized in a triplane representation. 
We drop the text conditioning.
The model comprises a VAE encoder-decoder pair, $(\vaeEncoder, \vaeDecoder)$, and a transformer-based denoising 
function, $\diffusionModelFn$.
We use the encoder $\vaeEncoder$ to encode the input image views, $\videoInput$, into a latent tensor, $\textbf{z}^{\videoInput}$.
Specifically, we encode $F=17$ views, evenly spaced in a 360-degree orbit around the object, alongside the six 
canonical views. 
The 17 views are encoded using the video VAE, and the canonical views are each encoded with the image VAE for 
additional quality~\citep{wang2024crm}.
We then run a single step of the denoising function, $\diffusionModelFn$, to produce an output latent, 
$\textbf{z}^{\trilat}(\theta) = \diffusionModelFn (\textbf{z}^{\videoInput})$. 
This latent 
is decoded via the VAE decoder, $\vaeDecoder$, into three triplane features:
$\left(\Txy, \Tyz, \Txz\right) = \vaeDecoder({\textbf{z}}^{\trilat}(\theta))$,
representing the $\mathrm{XY}$, $\mathrm{YZ}$, and $\mathrm{XZ}$ planes, respectively. 

We query the triplanes at each surface point, $\bp$, for each covered pixel in each view. 
These three point feature tensors are then concatenated to form a point feature, which is decoded by two 
small multi-layered perceptrons (MLPs) sharing the same input: a material decoder, $\mlpmat$, and an 
uncertainty decoder, $\mlpunc$.
By querying triplane features for all the surface points, 
the material and uncertainty decoders produce the respective material G-buffer, $\gbuffermat$, and per-material uncertainty (log-variance) G-buffer, $\gbufferunc$, for each covered pixel as follows: 
\begin{align}
\gbuffermat(\bp) &= \mlpmat([\Txy(\mathbf{p}), \Tyz(\mathbf{p}), \Txz(\mathbf{p})]), \\
\gbufferunc(\bp) &= \mlpunc([\Txy(\mathbf{p}), \Tyz(\mathbf{p}), \Txz(\mathbf{p})]).
\label{eq:uncertainty}
\end{align}
Note that we train over a large collection of 3D shapes, so both 
MLPs, $\mlpmat$ and $\mlpunc$, represent shared, \emph{universal} decoders of triplane features, while 
the predicted triplanes are unique per 3D object.

\paragraph{Training LMRM}
We jointly train the parameters of the LMRM network, $\theta$, and the decoder MLPs $\phi$ and $\psi$, 
by minimizing the training objective:
$\mathcal{L}(\theta, \phi, \psi) = \mathcal{L}_{\mathrm{mat}} + \lambda_{\mathrm{unc}}\mathcal{L}_{\mathrm{unc}}$,
which combines a material regression loss and an uncertainty loss.
The material loss supervises the mean prediction of $\mlpmat$ against the reference, \ie, ground-truth, G-buffers,
\begin{equation}
\mathcal{L}_{\mathrm{mat}} = \mathbb{E}_{\gbuffermat^{\mathrm{ref}}\sim\dataDistribution} \lVert \gbuffermat(\theta, \phi) - \gbuffermat^{\mathrm{ref}} \rVert_{2}^{2} ,
\label{eq:loss_mat}
\end{equation}
where $\gbuffermat^{\mathrm{ref}}$ denotes G-buffer views of the reference material parameters.
The uncertainty loss supervises the per-pixel material uncertainties
following the $\beta$-NLL formulation~\citep{Seitzer2022PitfallsOfUncertainty}:
\begin{equation}
\label{eq:loss_unc}
\mathcal{L}_{\mathrm{unc}}=\mathbb{E}\left[\lfloor \exp{(\beta\gbufferunc)} \rfloor \left( \frac{\gbufferunc}{2} {+} \frac{\lVert \lfloor \gbuffermat \rfloor - \gbuffermat^{\mathrm{ref}}\rVert_{2}^2}{2\exp(\gbufferunc)} \right) \right],
\end{equation}
where $\beta=0.5$ and $\lfloor\cdot\rfloor$ denotes the stop-gradient operation, applied per material channel.
We apply the stop-gradient to the material mean $\gbuffermat$ inside $\mathcal{L}_{\mathrm{unc}}$ so that it 
trains only the uncertainty decoder $\mlpunc$, while $\mathcal{L}_{\mathrm{mat}}$ trains the mean; please refer 
to~\citet{Seitzer2022PitfallsOfUncertainty} for details.

We adopt a two-stage training curriculum. We first pre-train LMRM on large-scale 3D shape datasets with 
annotated PBR materials~\citep{deitke2023objaverse,litman2025materialfusion,zhang2025texverse}, where 
$\mlpmat$ predicts the standard PBR G-buffers. 
This PBR pre-training stage lets the model learn a robust image-to-material mapping before facing the more 
expressive neural material target.
We then fine-tune LMRM on 3D assets annotated with procedurally enhanced neural materials~\citep{yu2026toward}, 
where $\mlpmat$ predicts the neural material G-buffer, \ie, diffuse base color $\rho_d$ and the specular 
latent code $\nmLatent$, exposing $\mlpunc$ to the increased ambiguity of recovering multi-lobe reflectance.

\section{Test-Time Optimization of Neural Materials}
\label{sec:tto}

\begin{figure}[t]
    \centering
    \begin{minipage}[c]{0.48\textwidth}
        \centering
        \includegraphics[width=\linewidth]{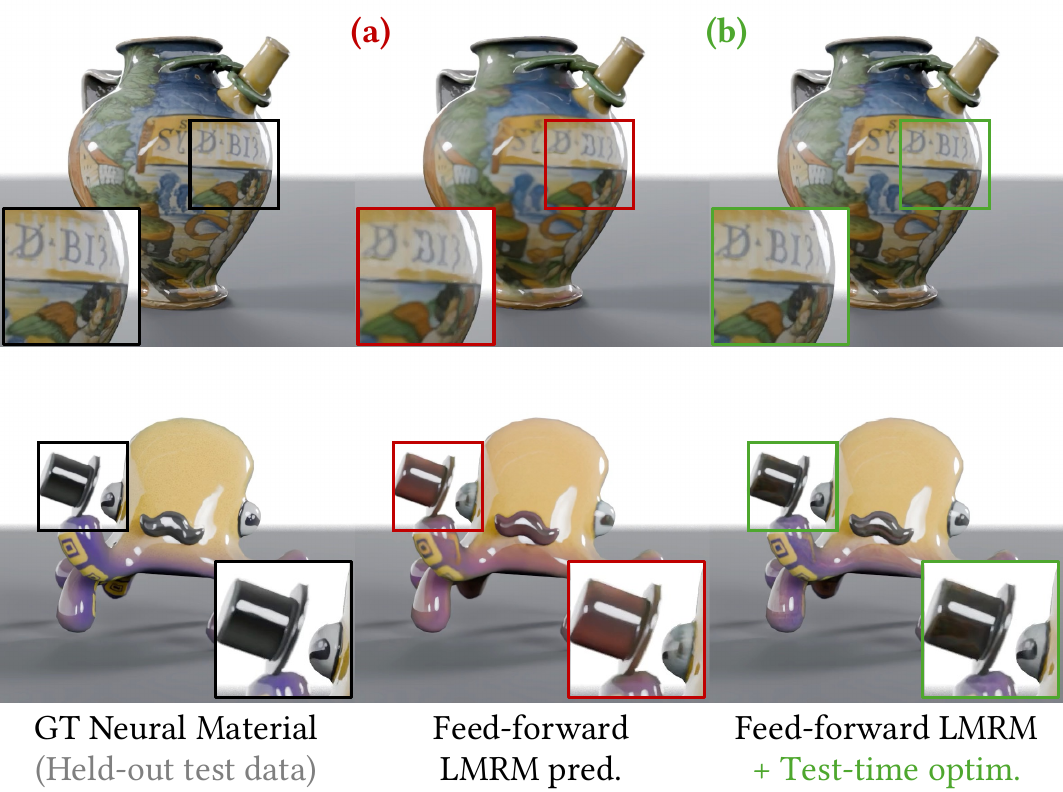}
        \vspace{-3mm}
        \caption{
        \textbf{Why need test-time optimization?} 
        While our feed-forward prediction gives reasonable initial materials (a), 
        test-time optimization (b) recovers finer details and corrects color shifts and material decomposition in
        challenging cases. All results relit with novel light probe.
        }
        \vspace{-1mm}
        \label{fig:why_tto}
    \end{minipage}
    \hspace{1mm}
    \begin{minipage}[c]{0.5\textwidth}
        \centering
        \includegraphics[width=\linewidth]{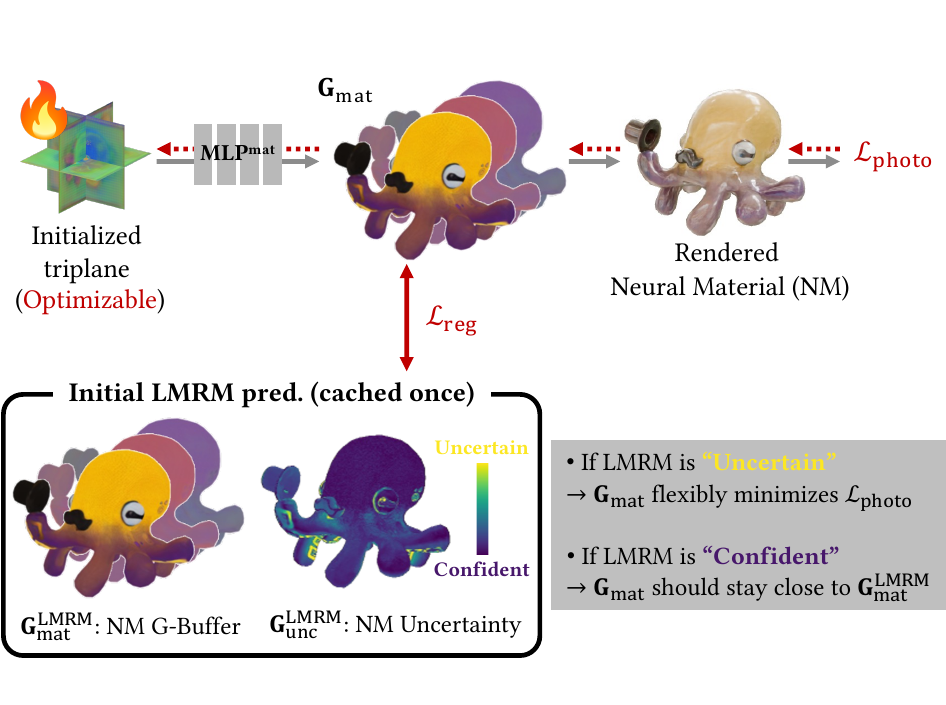}
        \caption{
        \textbf{Uncertainty-guided test-time optimization.}
        We optimize the LMRM-initialized triplane, decoding it into neural materials 
        and rendering via Monte Carlo path tracing (\emph{top}). An uncertainty-based 
        regularization anchors the solution to the initialization in low-uncertainty 
        regions (\emph{bottom}). 
        }
        \label{fig:unc_reg}
    \end{minipage}
\end{figure}

Given initial neural material predictions from LMRM, our test-time optimization (TTO) performs 
differentiable Monte Carlo inverse rendering to further refine material details and improve 
decomposition (see \Fref{fig:why_tto}).
Our MC inverse renderer follows NVDiffRecMC~\citep{hasselgren2022nvdiffrecmc} but with the PBR specular 
lobe replaced by a universal neural material basis~\citep{yu2026toward}.

\paragraph{Inverse rendering optimization for NM} 
In our setup, the neural component representing non-diffuse contributions consists of a universal, 
frozen decoder MLP $\nmDecoder$ operating on a $6$D latent code $\nmLatent(\bp)$ (\Eref{eq:nm_fwd}). 
Rather than optimizing the per-point material attributes directly, we expose them through our 
triplane representation $\bT$ (see \Sref{sec:overview}).
Together with the frozen material decoder $\mlpmat$, the triplane yields the per-point base color 
$\rho_d(\bp)$ of the traditional diffuse lobe~\citep{hasselgren2022nvdiffrecmc} and the specular 
latent $\nmLatent(\bp)$, \ie,
$\left(\rho_d(\bp),\, \nmLatent(\bp)\right) = \mlpmat(\bT(\bp))$,
so that the LMRM-initialized triplane $\bT$ (\Sref{sec:lmrm}) forms the only optimization parameter.

For a given camera pose $c$, our differentiable path tracer produces an image $\bI(\bT; c)$.
We light the scene with a known high dynamic range (HDR) environment map, and the renderer produces 
an output image with linear HDR values. We emphasize that our neural materials are operating 
in linear HDR; thus, strong specular peaks can have significant energy.
The reference image $\bI_\mathrm{ref}(c)$ from the same camera is stored in LDR.
We apply a tonemap operator $\calT$ with a learnable camera response function (CRF)~\citep{lin2025iris} that maps the renderer output to LDR.
The rendering loss is:
\begin{equation}
    \mathcal{L}_\mathrm{photo} = \E_{c}\Big[\lVert\calT\big(\bI(\bT; c)\big) - \bI_\mathrm{ref}(c)\rVert_{2}^{2}\Big].
    \label{eq:obj_photo}
\end{equation}
We minimize the rendering loss using Adam~\citep{Kingma2014}, with gradients 
obtained through differentiable path tracing.
Note that both the triplane material decoder $\mlpmat$ and the neural material decoder $\nmDecoder$ 
remain frozen throughout; only the triplane $\bT$ and the CRF parameters are updated.

\paragraph{Uncertainty-guided material regularization}
The photometric objective in \Eref{eq:obj_photo} is under-constrained at the neural material 
level: due to the rich multi-lobe specular effects, many combinations of base color $\rho_d$ 
and specular latent $\nmLatent$ can explain the same image-space observations. Thus, the 
test-time optimization may drift toward an implausible material while still minimizing the 
photometric loss.
To mitigate this ambiguity, 
we leverage the joint material-uncertainty prediction from 
LMRM (\Eref{eq:uncertainty}).
In a forward pass, LMRM predicts both the initial material mean $\gbuffermat^\text{LMRM}(\bp)$ 
and the log-variance $\gbufferunc^\text{LMRM}(\bp)$.
We freeze both as constants during TTO and use them to \emph{anchor} the current material 
$\gbuffermat(\bT)$, 
through an uncertainty-weighted regularization term:
\begin{equation}
    \mathcal{L}_\mathrm{reg} = \E_{\bp}\!\left[\,\frac{\lVert\gbuffermat(\bT; \bp) - \gbuffermat^\text{LMRM}(\bp)\rVert_2^2}{\exp\!\big(\gbufferunc^\text{LMRM}(\bp)\big)}\,\right],
    \label{eq:obj_reg}
\end{equation}
where the error is computed per material channel and reciprocally weighted by the
predicted uncertainty $\exp(\gbufferunc^\text{LMRM})$, mirroring the $\beta$-NLL term used in training (\Eref{eq:loss_unc}).
Intuitively, \Eref{eq:obj_reg} acts as a Gaussian prior centered at the LMRM prediction: 
confident 3D regions with small $\exp(\gbufferunc^\text{LMRM})$ are strongly anchored, 
suppressing material drift, while uncertain regions with large $\exp(\gbufferunc^\text{LMRM})$
are loosely constrained and free to drift from the feed-forward initialization (see \Fref{fig:unc_reg}).

Overall, the complete test-time optimization objective combines the photometric data term 
(\Eref{eq:obj_photo}) with the uncertainty-guided material regularizer (\Eref{eq:obj_reg}), \ie,
$\argmin_{\bT}\ \mathcal{L}_\mathrm{photo} + \lambda_\mathrm{reg}\,\mathcal{L}_\mathrm{reg}$,
where $\lambda_\mathrm{reg}$ weights uncertainty-guided material regularization for 
robust material decomposition.

\section{Experiments}
\label{sec:exp}

\begin{figure*}[t]
\centering
\includegraphics[width=\linewidth]{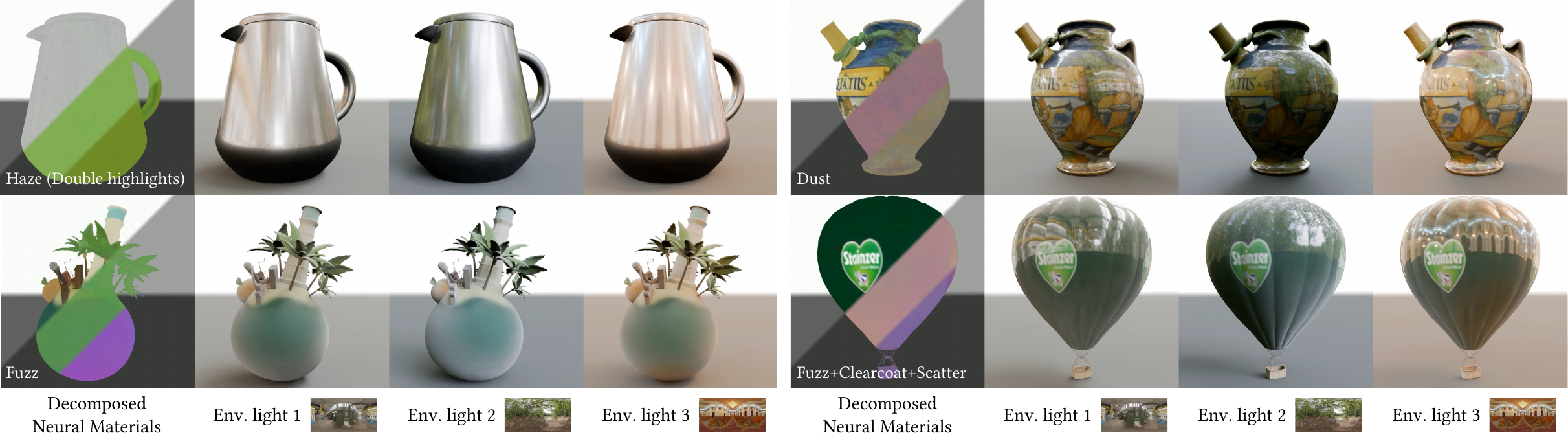}\vspace{-1.5mm}
\caption{
\ours reconstructs relightable neural materials from images, demonstrating complex multi-lobe effects, \eg, haze, clearcoat, dust, fuzz, scattering. More visual results in Figs.~\ref{fig:supp_relighting_1} \&~\ref{fig:supp_relighting_2}.
}
\label{fig:qual_ours}
\end{figure*}

In our experiments, we visualize renderings of neural materials extracted by \ours (\Fref{fig:qual_ours}) and 
compare with PBR material extraction methods on synthetic and real images. We also investigate the impact of design choices. Please refer to the appendix and attached visualizer for more results.

\begin{figure*}[t]
\centering
\includegraphics[width=\linewidth]{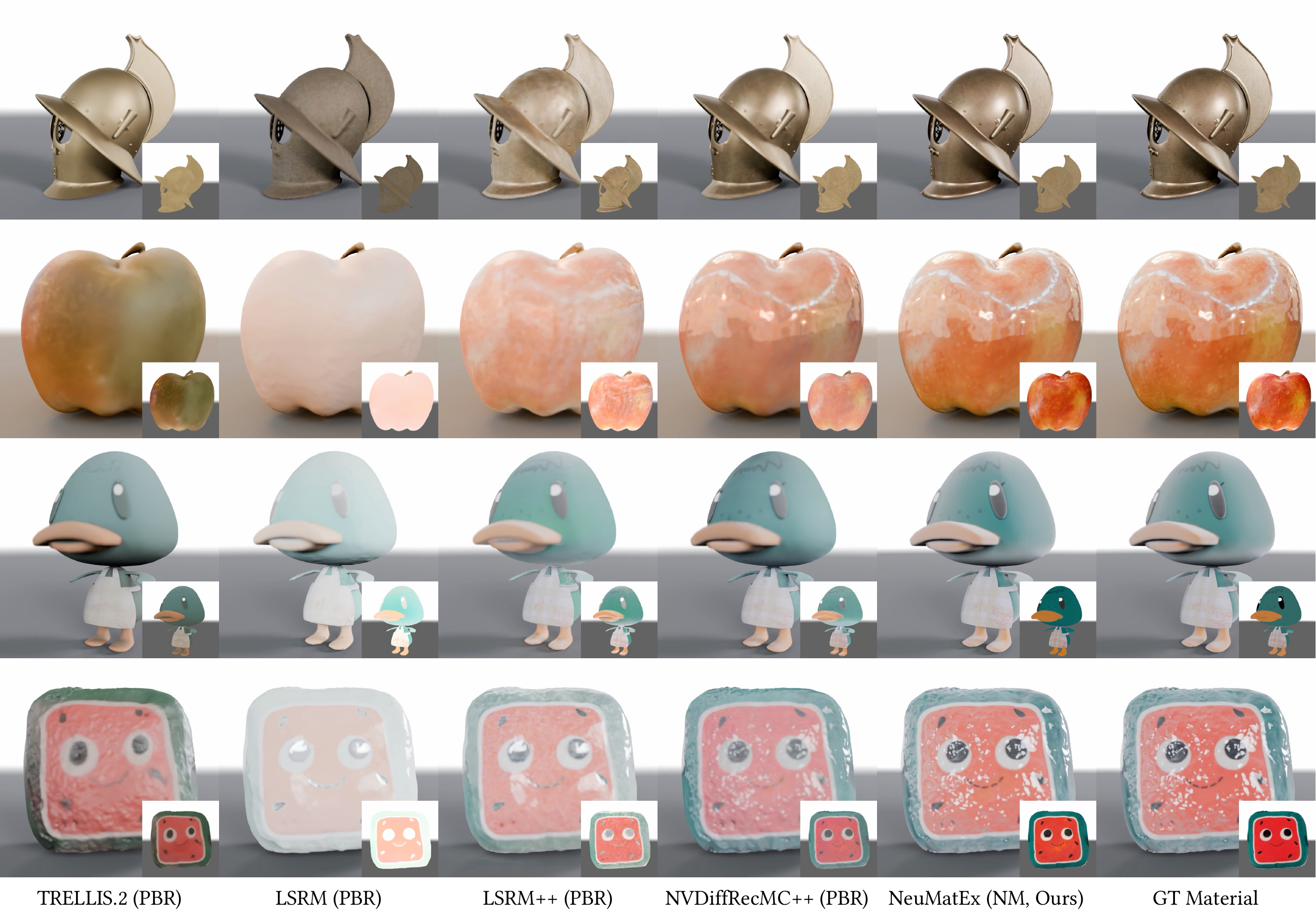}\vspace{-1.5mm}
\caption{\textbf{Qualitative: Material Extraction.} 
All methods use known fixed geometry except LSRM~\citep{li2026lsrm} and LSRM++, which predict their own; optimization methods use known poses and lighting.
TRELLIS.2~\citep{xiang2026trellis2} is a monocular feed-forward PBR model, included to highlight PBR limitations rather than for a fair comparison.
PBR-based methods fail to represent complex SVBSDFs, baking specular effects into the base color (see insets), while \ours decomposes them faithfully in the neural material latent space, yielding a clean base color.
}
\label{fig:qual_comparison}
\end{figure*}

\subsection{Comparison with PBR Material Extraction Methods}
For feed-forward PBR material extraction methods, we choose
Hunyuan3D-2.1~\citep{hunyuan3d21_2025_hunyuan3d} and TRELLIS.2~\citep{xiang2026trellis2}, using their texture reconstruction mode with known geometry.
Note that Hunyuan3D-2.1 and TRELLIS.2 are conditioned on a \emph{single} input view while our method uses multiple views;
this comparison therefore highlights PBR's limitations rather than being a direct comparison.
For a multi-view feed-forward method, we use LSRM~\citep{li2026lsrm}, which predicts PBR material and geometry, 
and further refine its PBR prediction with test-time optimization, denoted LSRM++.
We also extend \citet{hasselgren2022nvdiffrecmc} by initializing it with a PBR prediction 
from our PBR LMRM and optimizing only the materials, given known geometry and lighting. We denote this as NVDiffRecMC++. 
Both LSRM++ and NVDiffRecMC++ combine multi-view feed-forward PBR initialization with test-time optimization.
Finally, \ours with LSRM geometry examines whether our method remains effective once the known geometry assumption is relaxed.
We do not consider VideoNeuMat~\citep{xue2026videoneumat}: it targets planar 2D materials and no code is available.

\begin{table*}[t]
    \centering
    \caption{
        \textbf{Quantitative: Material Extraction.} For 40 test meshes with GT neural materials from~\citet{yu2026toward}, we compare reconstruction metrics for renderings (RGB) and base color (BC), averaged over 120 orbit views. Relighting metrics are averaged over four held-out HDR probes.
    }\vspace{-1.5mm}
    \label{tab:quant_comp}
    \resizebox{\textwidth}{!}{
        \begin{tabular}{l*{13}{c}}
        \toprule
        & \multicolumn{5}{c}{\textbf{Config.}}
        & \multicolumn{3}{c}{\textbf{Recon. RGB}}
        & \multicolumn{2}{c}{\textbf{Recon. Base color}}
        & \multicolumn{3}{c}{\textbf{Relighting}} \\

        \cmidrule(lr){2-6}
        \cmidrule(lr){7-11}
        \cmidrule(lr){12-14}

        \textbf{Method}
        & {Material}
        & {View}
        & {GT Mesh}
        & {Feed-fwd}
        & {TTO}
        & PSNR$\uparrow$
        & LPIPS$\downarrow$
        & SSIM$\uparrow$
        & siPSNR$\uparrow$
        & PSNR$\uparrow$
        & PSNR$\uparrow$
        & LPIPS$\downarrow$
        & SSIM$\uparrow$ \\
        
        \cmidrule{1-14}

        \textsc{Hunyuan3D-2.1}
        & \multirow{5}{*}{PBR}
        & Single
        & \greencheck
        & \greencheck
        & \redcross
        & 24.42
        & 0.137
        & 0.900
        & 24.41
        & 23.01
        & 22.50
        & 0.138
        & 0.902 \\

        \textsc{TRELLIS.2}
        & 
        & Single
        & \greencheck
        & \greencheck
        & \redcross
        & 23.55
        & 0.140
        & 0.912
        & 24.60
        & 23.95
        & 23.91
        & 0.138
        & 0.914 \\

        \textsc{LSRM}
        &
        & Multi
        & \redcross
        & \greencheck
        & \redcross
        & 18.29
        & 0.190
        & 0.870
        & 22.15
        & 12.54
        & 18.33
        & 0.190
        & 0.873 \\

        \textsc{LSRM++}
        &
        & Multi
        & \redcross
        & \greencheck
        & \greencheck
        & 20.68
        & 0.151
        & 0.891
        & 23.21
        & 16.41
        & 20.58
        & 0.155
        & 0.889 \\

        \textsc{NVDiffRecMC++}
        &
        & Multi
        & \greencheck
        & \greencheck
        & \greencheck
        & \cellcolor{tabsecond}{26.25}
        & \cellcolor{tabthird}{0.097}
        & \cellcolor{tabsecond}{0.938}
        & \cellcolor{tabthird}{26.31}
        & \cellcolor{tabthird}{24.89}
        & \cellcolor{tabsecond}{27.50}
        & \cellcolor{tabthird}{0.097}
        & \cellcolor{tabsecond}{0.937} \\

        \cmidrule{1-14}

        \textsc{\ours} (Ours, LSRM geom.)
        & \multirow{2}{*}{NM}
        & Multi
        & \redcross
        & \greencheck
        & \greencheck
        & \cellcolor{tabthird}{24.48}
        & \cellcolor{tabsecond}{0.093}
        & \cellcolor{tabthird}{0.921}
        & \cellcolor{tabsecond}{27.09}
        & \cellcolor{tabfirst}{26.42}
        & \cellcolor{tabthird}{26.23}
        & \cellcolor{tabsecond}{0.092}
        & \cellcolor{tabthird}{0.923} \\

        \textsc{\ours} (Ours, Full)
        & 
        & Multi
        & \greencheck
        & \greencheck
        & \greencheck
        & \cellcolor{tabfirst}{34.78}
        & \cellcolor{tabfirst}{0.050}
        & \cellcolor{tabfirst}{0.963}
        & \cellcolor{tabfirst}{31.45}
        & \cellcolor{tabsecond}{25.30}
        & \cellcolor{tabfirst}{33.13}
        & \cellcolor{tabfirst}{0.047}
        & \cellcolor{tabfirst}{0.966} \\

        \bottomrule
        \end{tabular}
    }
\end{table*}

In \Fref{fig:qual_comparison}, we show renderings of extracted materials alongside the diffuse base 
color for assets with complex specular effects.
We use ground-truth neural materials in the basis of \citet{yu2026toward} as a reference; all methods are 
rendered from the input view of TRELLIS.2 for a fair comparison.
\ours reconstructs materials close to the reference, capturing the specular effects and separating 
the diffuse and specular terms without baking specular shading into the base color. 
In \Tref{tab:quant_comp}, we report reconstruction scores over 120 orbit views for both the path-traced renderings and base color, as well as relighting metrics under four held-out HDR probes. For base color, we additionally report scale-invariant PSNR (siPSNR), 
which factors out the inherently ambiguous global intensity scale of the recovered base color.
\ours outperforms all PBR-based methods on both, showing that the expressiveness of 
neural materials is essential for appearance effects PBR cannot capture.
Adding the same test-time optimization to the PBR baselines (LSRM++, NVDiffRecMC++) narrows but does 
not close this gap, indicating the limitation lies in the representation rather than the optimization.
Finally, even with the imperfect predicted mesh, \ours extracts materials faithful enough to outperform the competing methods (LSRM, LSRM++) under the same predicted geometry by a large margin, showing that \ours generalizes beyond the known geometry setting.

\begin{table}[tb]
    \caption{
        \textbf{Effect of the \ours design choices.}
        We ablate material parameterization (Param.), LMRM initialization (LMRM init.), and 
        uncertainty regularization (Unc. reg.) on 40 held-out test meshes from~\citet{yu2026toward}.
        Our full method (f) achieves the best material decomposition and competitive render 
        quality, leveraging a large-scale trained prior and uncertainty-guided regularization.
    }
    \centering
    \resizebox{0.65\linewidth}{!}{%
        \begin{tabular}{cccc ccc}
            \toprule
            \multicolumn{4}{c}{\textbf{\ours Config.}}
                & \multicolumn{2}{c}{\textbf{Material Decomp.} ($\uparrow$)}
                & \multicolumn{1}{c}{\textbf{Render Quality} ($\uparrow$)} \\
            \cmidrule(lr){1-4} \cmidrule(lr){5-6} \cmidrule(l){7-7}
            & Param. & LMRM init. & Unc. reg. & PSNR\textsubscript{BaseColor} & PSNR\textsubscript{Latents} & PSNR\textsubscript{PathTrace} \\
            \cmidrule{1-7}
            (a) & \multirow{3}{*}{UV} & \redcross & \redcross & 12.19\std{2.70}             & 14.74\std{1.31}             & 31.67\std{1.79} \\
            (b) &                     & \greencheck & \redcross & 22.73\std{3.71}             & 25.40\std{3.67}             & \cellcolor{tabfirst}{35.67\std{1.80}}  \\
            (c) &                     & \greencheck & \greencheck & \cellcolor{tabthird}{23.67\std{3.16}}             & \cellcolor{tabthird}{26.32\std{4.17}}             & 33.94\std{2.57}   \\
            \cmidrule{1-7}
            (d) & \multirow{3}{*}{Triplane} & \redcross & \redcross & 13.56\std{3.50}             & 17.15\std{2.56}             & 28.67\std{3.03} \\
            (e) &                           & \greencheck & \redcross & \cellcolor{tabsecond}{24.84\std{3.97}} & \cellcolor{tabsecond}{26.48\std{4.42}} & \cellcolor{tabsecond}{35.51\std{1.93}} \\
            (f) &                           & \greencheck & \greencheck & \cellcolor{tabfirst}{25.30\std{3.93}}    & \cellcolor{tabfirst}{26.81\std{4.45}}    & \cellcolor{tabthird}{34.78\std{2.22}} \\
            \bottomrule
        \end{tabular}%
    }
    \label{tab:ablation}
\end{table}

\subsection{Ablations}
\paragraph{Impact of design choices} 
In \Tref{tab:ablation}, we ablate the design choices of \ours: the triplane 
material representation, the impact of the initial material prediction from the 
LMRM, and the uncertainty regularization.
Neural materials are significantly harder to extract from multi-view images through 
inverse rendering, and we note that the strong prior from the LMRM initialization 
greatly improves both the render quality and the material decomposition compared 
to optimizing from randomly initialized materials ((a),(d) vs. (b),(e)). 
Uncertainty-guided optimization has a more subtle impact but increases 
the quality of the material decomposition ((b),(e) vs. (c),(f)). 
The triplane representation provides a significant quality improvement, since 
it enforces stricter spatial coherence than na\"ive UV texture mapping, which 
helps regularize test-time optimization ((a)-(c) vs. (d)-(f)).

\paragraph{Impact of uncertainty-guided regularization}
In \Tref{tab:ablation_tto_reg}, we ablate the benefit of uncertainty-guided 
regularization for test-time optimization. 
The unconstrained TTO ({row} (b)) already outperforms the LMRM prediction alone, 
and adding uncertainty-guided regularization further improves material 
decomposition while retaining competitive render quality ({row} (c)).
Also, we evaluate two regularization strategies using our predicted uncertainty 
$\gbufferunc^\text{LMRM}$, and an oracle $\gbufferunc^\text{oracle}$ that uses 
material G-buffer errors as uncertainty, \ie, an ideal upper bound, up to scale ({row} (d)).
Our TTO with predicted $\gbufferunc^\text{LMRM}$ performs close to the ideal case, 
supporting that LMRM produces well-calibrated uncertainty estimates that effectively 
guide TTO.
In \Fref{fig:ablation_regularizer}, we qualitatively show a clear improvement 
in material decomposition from uncertainty-guided regularization.

\begin{table}[t]
    \vspace{-1.5mm}
    \caption{
        \textbf{Ablation on TTO regularization.} 
        We compare TTO results with and without uncertainty-guided regularization,
        evaluated on 40 held-out test meshes from~\citet{yu2026toward}. Our predicted 
        uncertainty (c) closely approaches the oracle upper bound (d), confirming 
        the predicted uncertainty effectively guides TTO. Please refer to visual ablation results in \Fref{fig:ablation_regularizer}.
    }\vspace{-1.5mm}
    \centering
    \resizebox{0.65\linewidth}{!}{
    \begin{tabular}{l ccc}
        \toprule
            & \multicolumn{2}{c}{\textbf{Material Decomp. ($\uparrow$)}} & \multicolumn{1}{c}{\textbf{Render Quality ($\uparrow$)}} \\
            \cmidrule{2-3} \cmidrule{4-4}
            \textbf{TTO Regularization Config.} & PSNR\textsubscript{BaseColor} & PSNR\textsubscript{Latents} & PSNR\textsubscript{PathTrace}\\
            \cmidrule{1-4}
            (a)\,\,No TTO (Only LMRM pred.)                              & 23.82\std{3.72}            & 26.21\std{4.03}            & 31.99\std{2.95} \\
            (b)\,\,TTO \emph{w/o} regularization                              & \cellcolor{tabsecond}{24.84\std{3.97}} & \cellcolor{tabsecond}{26.48\std{4.42}} & \cellcolor{tabfirst}{35.51\std{1.93}} \\
            (c)\,\,TTO \emph{w/} $\gbufferunc^\text{LMRM}$ (Ours)   & \cellcolor{tabfirst}{25.30\std{3.93}} & \cellcolor{tabfirst}{26.81\std{4.45}}    & \cellcolor{tabsecond}{34.78\std{2.22}} \\
            \cmidrule{1-4}
            (d) TTO \emph{w/} $\gbufferunc^\text{oracle}$ (Ideal upper bound) & \cellcolor{Gray}{25.99\std{4.39}} & \cellcolor{Gray}{27.31\std{4.44}} & \cellcolor{Gray}{35.05\std{2.17}} \\
        \bottomrule
    \end{tabular}
    }
    \label{tab:ablation_tto_reg}
\end{table}

\begin{figure}[tb]
\centering
\includegraphics[width=\linewidth]{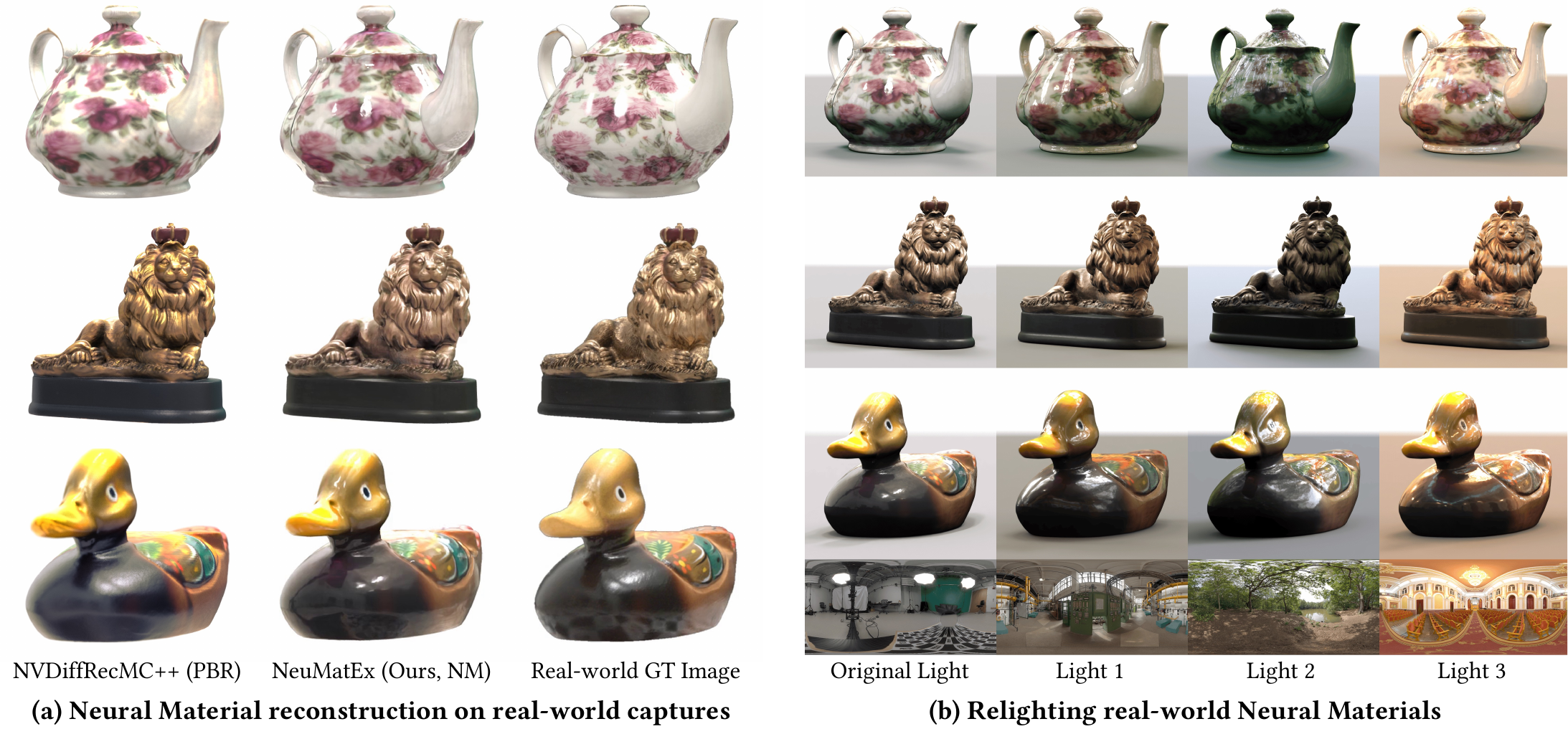}
\caption{\textbf{Neural materials from real-world captures.}
(a) We apply \ours to extract neural materials from real-world multi-view object capture dataset~\citep{dong2025dtc}. 
Our extracted materials capture effects beyond the standard PBR, \eg, clearcoat on the teapot. (b) Relit renderings of our extracted neural materials show that the recovered reflectance generalizes across illumination conditions, faithfully reproducing complex specular behavior without baked-in effects.}
\label{fig:realworld}
\end{figure}

\subsection{Neural Materials from Real-world Captures}
Real-world objects exhibit complex reflection properties that go beyond the 
expressiveness of the standard PBR model.
Extracting neural materials from real photos is a step toward data-driven SVBSDF authoring.
As a proof-of-concept, we apply {\ours} to extract neural materials from real-world captures.
We use the Digital Twin Catalog (DTC)~\citep{dong2025dtc} consisting of high-quality 
multi-view captures,
meshes, camera poses, and environment lighting. 
We compare {\ours} against NVDiffRecMC++ in \Fref{fig:realworld}\colorref{(a)}. 
The PBR materials struggle with complex glossy effects and layered reflectance, 
while our extracted neural materials are expressive enough to capture them, 
producing more faithful reflections that remain stable under novel illumination 
(\Fref{fig:realworld}\colorref{(b)}).

\begin{figure}[t]
\centering
\includegraphics[width=0.86\linewidth]{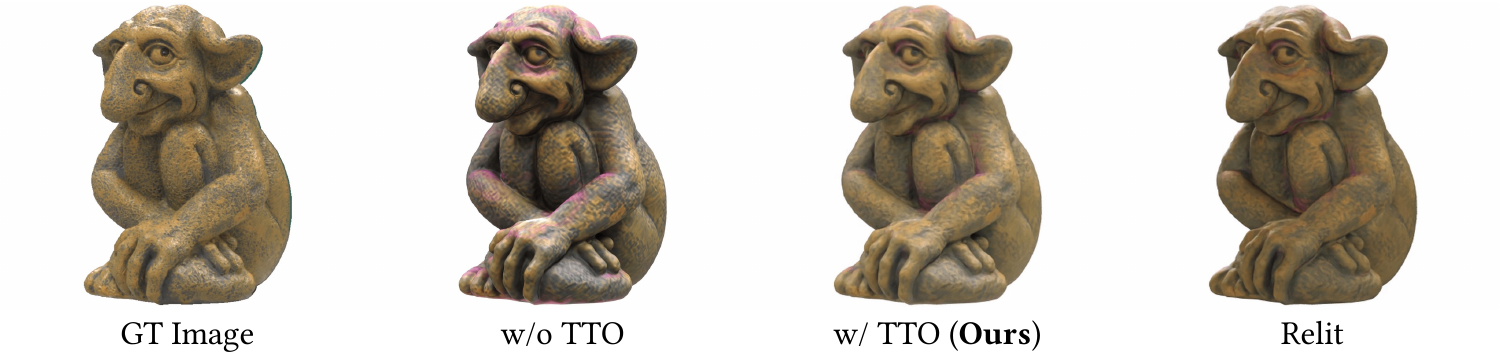}\vspace{-1.5mm}
\caption{
\textbf{Limitations.}
We build on a pre-trained neural material basis~\citep{yu2026toward}, inheriting its limitations, where if predicted materials lie outside the valid domain of the basis, specular artifacts can appear. 
Our test-time optimization mostly mitigates this, but artifacts may still be observed.
}\vspace{-1.5mm}
\label{fig:realworld:limitations}
\end{figure}

\section{Discussion and Limitations}
\label{sec:discussion}

We present \ours, demonstrating that inverse rendering can recover expressive neural materials from images. 
Combining a strong feed-forward initialization from the Large Material Reconstruction Model with uncertainty-guided regularization stabilizes ill-posed test-time optimization in the neural material latent space.
\ours outperforms PBR-based methods and recovers richer materials in synthetic and real-world scenarios, cleanly separating lighting from material response. 
While we build on the SVBSDF representation of \citet{yu2026toward}, our method can be extended to other neural material models. 
In future work we aim to improve LMRM material prediction quality and the resolution of the recovered material to capture finer appearance details.

\paragraph{Limitations} 
The complexity of the neural basis remains challenging even with careful initialization.
In \Fref{fig:realworld:limitations}, LMRM predictions on out-of-domain inputs 
such as real-world photographs can exhibit specular artifacts with a red tint, as also 
observed by~\citet{yu2026toward}. 
Test-time optimization often reduces this, but slight artifacts can remain. 
Learning a more robust neural material model, ideally from real-world observations, 
is an important direction toward photorealistic reflectance recovery.

\subsection*{Acknowledgments}

We thank Milo\v{s} Ha\v{s}an, Tizian Zeltner, Saeed Hadadan, and Yunchen Yu for helpful discussions 
and sharing code. We also thank Aaron Lefohn and Chris Wyman for supporting this research.
The teaser image (\Fref{fig:teaser}) uses the MaterialX Teapot and Lion assets \textcopyright{} 2026 NVIDIA Corporation
(ASWF Digital Assets License v1.1), available at \url{https://github.com/DigitalProductionExampleLibrary/MaterialXTeapotLion}.

\bibliography{main}
\bibliographystyle{iclr2027_conference}

\addtocontents{toc}{\protect\setcounter{tocdepth}{2}}
\newpage
\title{Extracting Neural Materials from Images\\\vspace{2.0mm} --- Supplementary material ---}

\makeatletter
  \let\maketitle\savedmaketitle \let\@maketitle\savedmk \let\@author\savedauthor
  \makeatother
\maketitle

\appendix

\newtheorem{assume}{Assumption}
\newtheorem{definition}{Definition}
\newtheorem{lemma}{Lemma}

\setcounter{section}{0}
\setcounter{figure}{0}
\setcounter{table}{0}
\setcounter{equation}{0}

\renewcommand\thesection{\Alph{section}}
\renewcommand\thefigure{S\arabic{figure}}
\renewcommand{\thetable}{S\arabic{table}}
\renewcommand\theequation{\alph{equation}}

\makeatletter
\renewcommand{\numberline}[1]{#1\hspace{1.0em}}
\makeatother

\vspace{-3mm}
\begingroup
\hypersetup{linkcolor=black}
\makeatletter
\renewcommand{\l@section}[2]{\addpenalty{-\@highpenalty}%
  \addvspace{2pt}\@dottedtocline{1}{0em}{1.5em}{\bfseries #1}{\bfseries #2}}
\renewcommand{\l@subsection}{\@dottedtocline{2}{1.2em}{2.3em}}
\makeatother
\noindent\rule{\linewidth}{1.0pt}
\vspace{-6mm}
\tableofcontents
\noindent\rule{\linewidth}{1.0pt}
\endgroup

\section{Implementation Details}

\subsection{Large Material Reconstruction Model}
\label{sec:impl_lmrm}
\paragraph{Multi-view input encoding} 
We encode $F=17$ views, evenly spaced in a 360-degree orbit around the object, alongside the six canonical views. The 17 views are encoded using the frozen Wan2.1 VAE encoder, and the canonical views are each encoded with the image VAE for additional quality~\citep{wang2024crm}.
The input views ($F$ orbital + six canonical) are encoded into a latent tensor, $\bz^{\videoInput}\in\mathbb{R}^{(1+(F-1)/4+6) \times 16 \times H/8 \times W/8}$. We use $H=W=512$ for the input image resolution.
We process all the orbital input views jointly using Wan2.1's video VAE encoder, allowing the subsequent LMRM backbone to aggregate cross-view information.

\paragraph{Backbone} The backbone architecture of LMRM is a diffusion transformer (DiT)~\citep{peebles2022dit}, initialized from the pre-trained text-to-video Wan2.1-1.3B~\citep{wan2025wan}, repurposed as a \emph{single-step} model.
The backbone performs a single deterministic forward pass at a fixed timestep $t{=}0$, rather than iterative denoising, so that the DiT acts as a feed-forward regressor that takes multi-view images and predicts triplane feature planes. 
As the reconstruction task is not text-conditioned, we bypass the text cross-attention by feeding a fixed null-text embedding to every layer. 
We keep the original Wan2.1-1.3B configuration: $30$ transformer blocks with model width $1536$, $12$ attention heads, feed-forward inner dimension $8960$, patch embedding ($1{\times}2{\times}2$), and 3D rotary position embeddings~\citep{su2021roformer}.

\paragraph{Triplane decoding head}
After the encoded input latent tensor, $\bz^{\videoInput}$, is processed by the DiT backbone, the output tokens are mapped to triplane feature planes through a lightweight linear head followed by a convolutional decoder initialized from the pre-trained Wan2.1 VAE decoder (kept trainable). 
We decode the latent frames and obtain three triplane features: $\Txy, \Tyz, \Txz$ representing the  $\mathrm{XY}$, $\mathrm{YZ}$, and $\mathrm{XZ}$ planes, respectively, with 12 channels each. 
We decode each feature plane independently to avoid the causal temporal coupling of the video VAE decoder.

\paragraph{Triplane MLPs}
Given a 3D surface point,
we bilinearly sample the three axis-aligned planes and concatenate the resulting $C{=}12$-channel features ($3C{=}36$ total). We further append a NeRF-style positional encoding~\citep{mildenhall2020nerf} ($3$ frequency bands) for the 3D surface position and decode the material parameters with a triplane MLP, \ie, a $4$-layer MLP of width $64$ with intermediate ReLU activations, and a final sigmoid (only for the material head). 
For the PBR pre-training stage, the material MLP head predicts $5$ channels $=$ base color (3D) + roughness (1D) + metallicity (1D).
For the neural material fine-tuning stage, the material MLP head predicts $9$ channels $=$ base color (3D) + neural specular latents (6D). 
A parallel uncertainty MLP head, implemented as a separate MLP of identical capacity, predicts a per-channel log-variance used in a $\beta$ negative log-likelihood ($\beta$-NLL) objective (\Eref{eq:loss_unc}).
The MLPs are evaluated only at surface points via a gather-scatter scheme.

\paragraph{Training}
We train the LMRM for $200$K steps on $32$ NVIDIA A100 GPUs with a per-GPU batch size of $2$ in \texttt{bfloat16}, using AdamW with a cosine learning rate scheduler. 
For learning rates, we use $5{\times}10^{-5}$ for the backbone, and $5{\times}10^{-4}$ for both the material and uncertainty triplane MLPs. 
The $\beta$-NLL loss is disabled for the first $30$K steps and then linearly ramped over $10$K steps. 
Losses are computed on eight randomly sampled supervision views per object.

\subsection{Test-time Optimization}\label{sec:impl_tto}
In this stage, we optimize only the LMRM-initialized triplanes (the triplane MLPs are kept frozen) using Adam with a learning rate of $5\times10^{-3}$.
During TTO, we use differentiable Monte Carlo path tracing with $128$ samples per pixel (spp).
For the uncertainty-guided material regularization, we cache the initial material predictions and per-channel aleatoric uncertainties from the frozen MLPs once before optimization and keep them fixed throughout.
We set $\lambda_\mathrm{reg}=2.0$, as this value was selected from a sweep over $\{0.1, 0.25, 0.5, 1, 2\}$ to best balance rendered-image and material-decomposition quality.
Test-time optimization takes $15$ minutes per object ($1000$ steps at $512\times512$ resolution) on a single NVIDIA A100 GPU, excluding final rendering.
We use the Blender AgX tonemapper for optimization and for all results reported in the paper, unless otherwise noted.

\subsection{Neural Material Path Tracer}\label{sec:impl_render}

For our path tracer with neural materials, the outgoing radiance $L(\omega_o, \bp)$ at a surface 
point $\bp$ in direction $\omega_o$ is defined by the rendering equation~\citep{Kajiya1986}:
\begin{equation}
	L(\omega_o, \bp) = \int_\Omega L_i(\omega_i, \bp)f(\bp, \omega_i,\omega_o) (\omega_i \cdot \mathbf{n}) d\omega_i.
    \label{eq:ibl}
\end{equation}
This is an integral of the product of the incident radiance $L_i(\omega_i, \bp)$ from direction 
$\omega_i$ and the neural material SVBSDF $f(\bp, \omega_i, \omega_o)$ (\Eref{eq:svbsdf}). The 
integration domain is the hemisphere $\Omega$ around the surface normal, $\mathbf{n}$. 
We evaluate the rendering equation using Monte Carlo integration:
\begin{equation}
	L(\omega_o,\bp) \approx \frac{1}{N} \sum_{i=1}^{N} \frac{L_i(\omega_i,\bp)f(\bp,\omega_i,\omega_o) (\omega_i \cdot \mathbf{n})}{p(\omega_i)},
\end{equation}
with samples drawn from some distribution $p(\omega_i)$.

We use \emph{multiple importance sampling}~\citep{Veach1995} (MIS), a framework for combining 
multiple sampling techniques to reduce variance in Monte Carlo integration. 
We apply MIS with two sampling techniques: light importance sampling, 
$p_\mathrm{light}(\omega)$, using sampling from a piecewise-constant 2D distribution~\citep{Pharr2023}, 
and BSDF sampling, $p_\mathrm{bsdf}(\omega)$. The weighting functions $w_k(x)$ for each technique 
are chosen using the \emph{balance heuristic}~\citep{Veach1995,Pharr2023}. 

Following~\citet{zeltner2024neural}, our BSDF sampler uses a frozen sampling decoder to 
extract parameters for two independent GGX lobes for the neural part, then selects one component among 
the diffuse cosine lobe and the two GGX lobes according to the predicted mixture weights (\Eref{eq:nm_fwd}) 
and samples it.
Using the taxonomy of differentiable Monte Carlo estimators~\citep{Zeltner2021}, our importance sampling 
is \emph{detached}, \ie, gradients are not back-propagated to scene parameters in the sampling step, 
only in the material evaluation.

\subsection{Neural Material Rendering Performance}\label{sec:impl_perf}
\begin{figure}[ht]
\centering
\includegraphics[width=\linewidth]{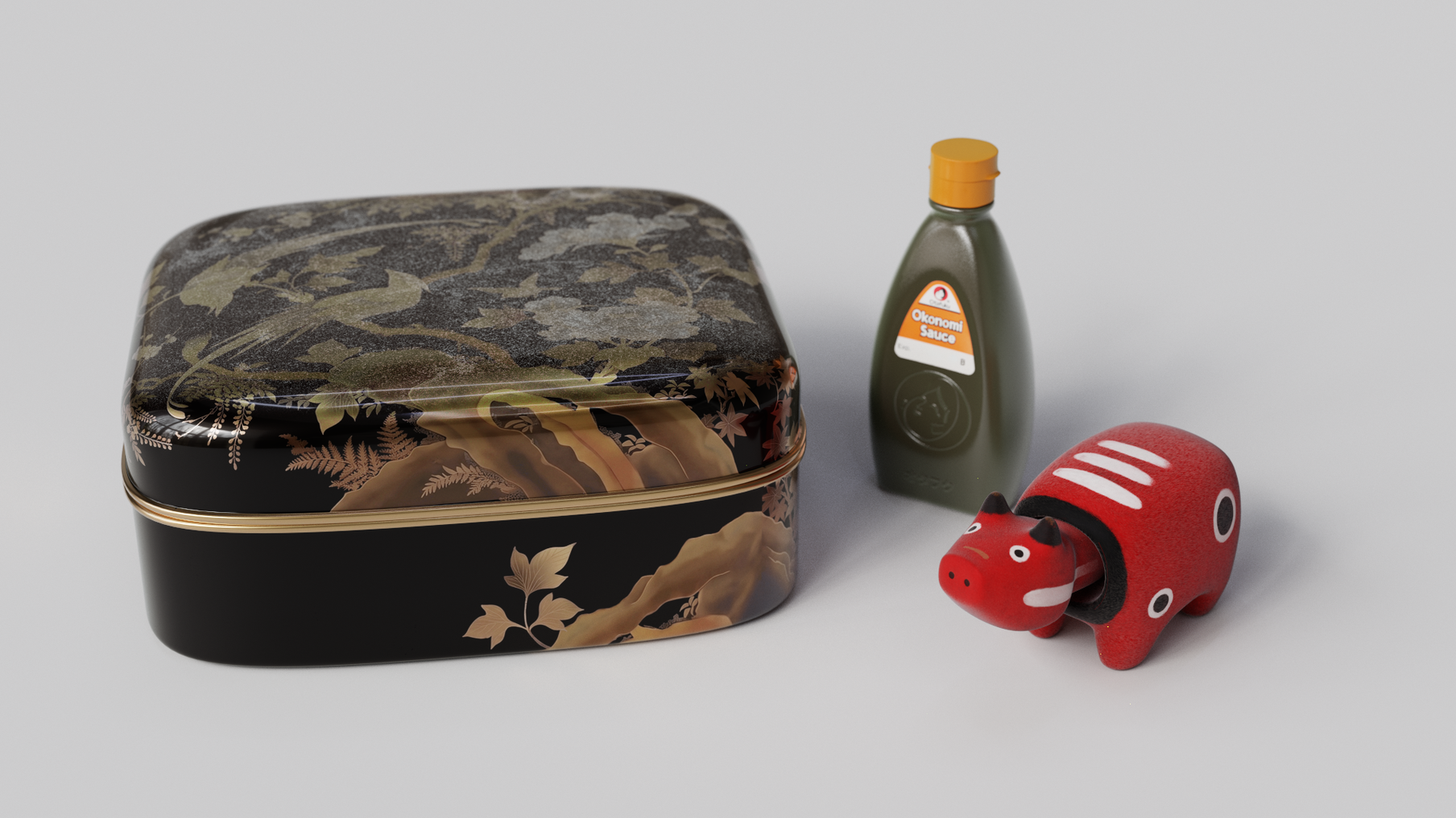}
\caption{
\textbf{Rendering performance}. For a set of reference neural materials, the runtime performance for path tracing in 1080p resolution with 1~spp and 10 bounces is ${\sim}4$~ms on an RTX 5090.   
}
\label{fig:runtime}
\end{figure}

We evaluate rendering performance of the neural materials at 1080p resolution 
on an RTX 5090 using the custom test scene in \Fref{fig:runtime}, which consists 
of four different neural materials and runs at 3.88~ms per frame at 1~spp in our path 
tracer with 10 bounces. 
Note that this number reflects the combined cost of both the 
neural evaluation and the underlying path-tracing computation. 
The universal neural material basis consists of MLPs with 4 layers of width 64 (4×64) 
for BSDF evaluation, a 3×64 auxiliary network for estimating transmission albedo, and 
a 2×32 sampling network for importance sampling.
This clearly indicates that the neural material representation we extract allows for 
real-time performance within a path-tracing context.
Please refer to~\citet{zeltner2024neural} for a thorough analysis of 
neural material runtime performance and implementation details.

\subsection{\ours for Real-world Photos}\label{sec:impl_realworld}

We use the Digital Twin Catalog (DTC)~\citep{dong2025dtc} consisting of high-quality 
multi-view captures ($\sim120$ photographs per object) together with meshes, camera 
poses, and measured environment lighting. 
First, we use novel-view synthesis to obtain the preset views, \ie, $17$ orbital 
views and $6$ orthogonal views, needed for our LMRM. 
Specifically, we use 2D Gaussian Splatting~\citep{Huang2DGS2024}, where we initialize 
the primitives at the mesh surface and the normals with the mesh normals.
Then, we run our LMRM on the rendered views to estimate the initial triplane features for neural materials.
During the test-time optimization, we optimize against all real photograph views to avoid baking any novel-view-synthesis artifacts into the materials.

For all of these experiments, we use the captured mesh and HDR environment map, together with the posed LDR images. 
A constant tonemapping operator for the rendered images during our test-time optimization can introduce a domain gap with the real-world observations.
To avoid this, we follow IRIS~\citep{lin2025iris} and use a learnable camera response function (CRF) for these real-world experiments.
We use the parametric CRF space of IRIS~\citep{lin2025iris}, based on Grossberg and Nayar's empirical model of response (EMoR)~\citep{grossberg04emor}, which fits a PCA basis to real-world measured CRFs. 
The optimized CRFs are then used to tonemap the rendered images for the comparisons as well as for the relightings.

\section{More Results}

\subsection{Qualitative Ablation on Uncertainty-guided Regularization}

\begin{figure}[h]
\centering
\includegraphics[width=\linewidth]{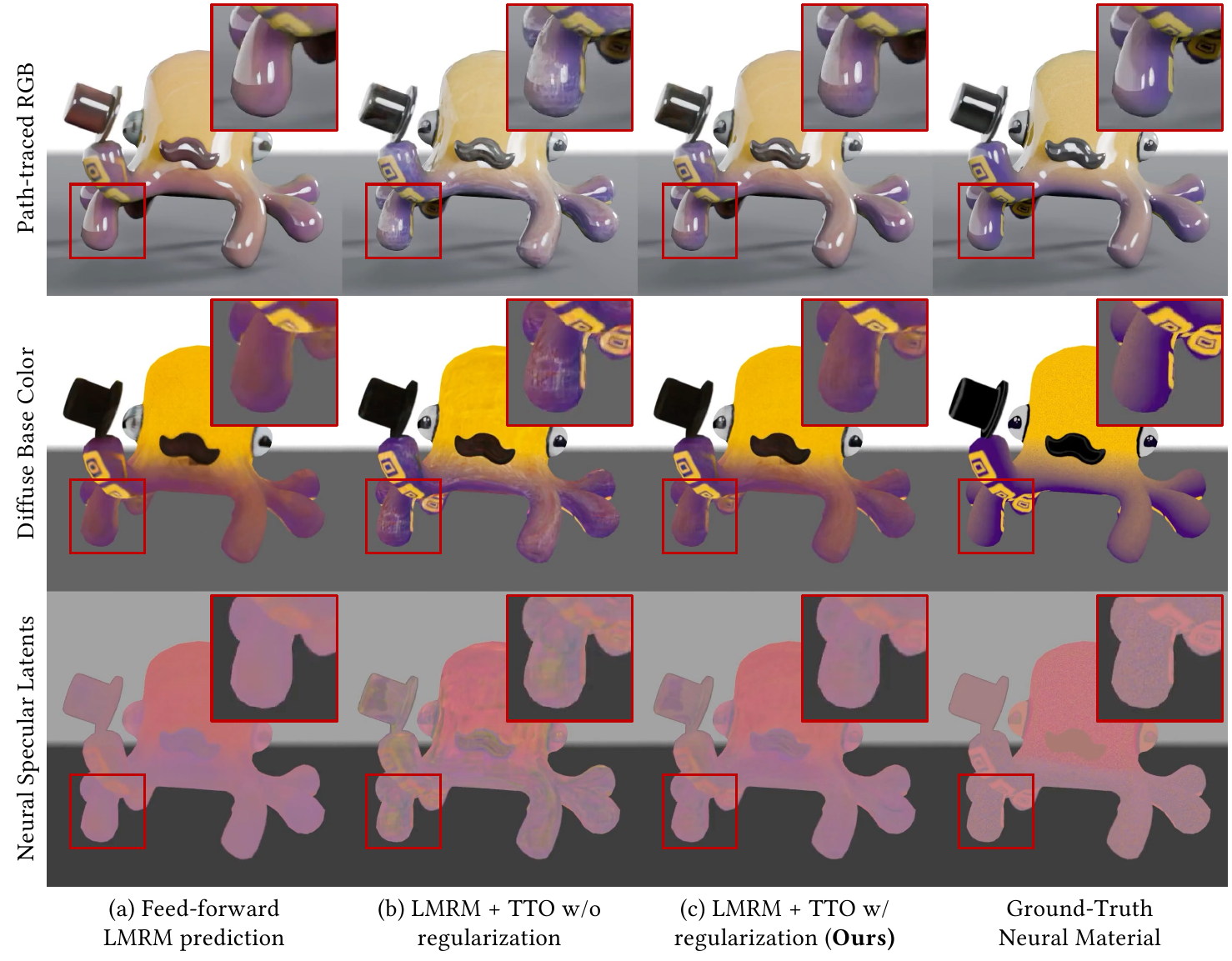}
\caption{
\textbf{Effects of uncertainty-guided regularization.}
Unconstrained TTO (column 2) suffers from lighting baked into base color and neural latents.
Our uncertainty-guided regularization (column 3) yields well-behaved and detailed material intrinsics
over the LMRM initialization (see crops).
}
\label{fig:ablation_regularizer}
\end{figure}

Figure~\ref{fig:ablation_regularizer} provides the qualitative ablation results for the proposed uncertainty-guided regularization in the test-time optimization.
All configurations reproduce the path-traced renderings faithfully (first row), 
as the 
rendering objective alone can be satisfied by many material configurations.
The difference instead appears in the decomposed material intrinsics.
Our uncertainty-guided regularization anchors the optimization toward the LMRM 
prediction where it is confident, and relaxes it only where the predicted uncertainty is high.
This keeps the base color and the specular latents free of baked-in lighting, while still recovering local details that the feed-forward prediction misses.

\clearpage

\subsection{Relighting Neural Materials Extracted by \ours}

\begin{figure*}[htbp]
\centering
\includegraphics[width=\linewidth]{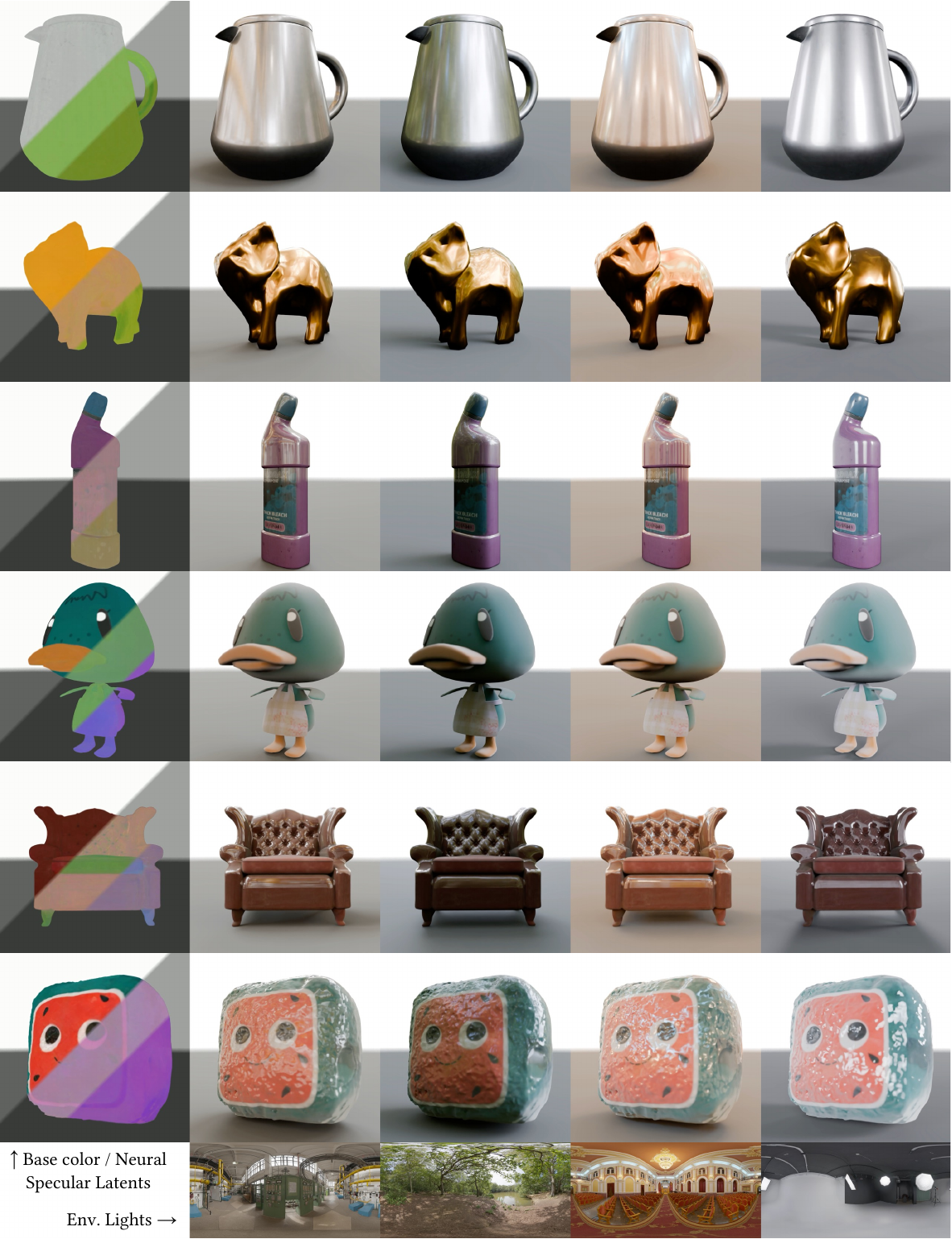}
\caption{\textbf{Visualization of the extracted neural materials.} Column 1 shows the decomposed neural materials (diffuse base color + neural specular latents A \& B), and columns 2-5 show the 128spp relit renderings using four different HDR light probes.
}
\label{fig:supp_relighting_1}
\end{figure*}

\begin{figure*}[t]
\centering
\includegraphics[width=\linewidth]{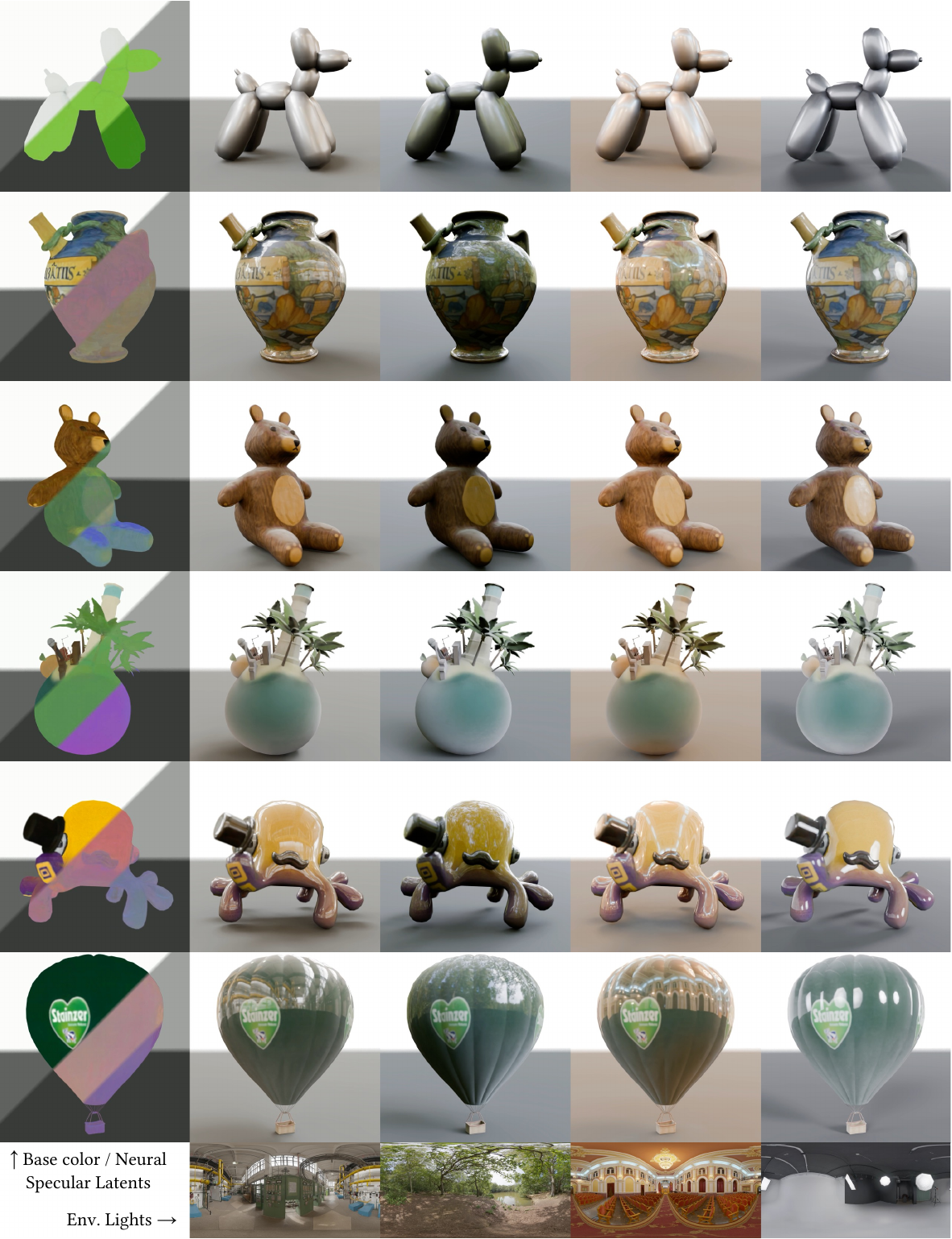}
\caption{\textbf{Visualization of the extracted neural materials.} Column 1 shows the decomposed neural materials (diffuse base color + neural specular latents A \& B), and columns 2-5 show the 128spp relit renderings using four different HDR light probes.
}
\label{fig:supp_relighting_2}
\end{figure*}

\clearpage

\subsection{More Comparisons with PBR Material Extraction Methods}
\begin{figure*}[ht]
\centering
\includegraphics[width=\linewidth]{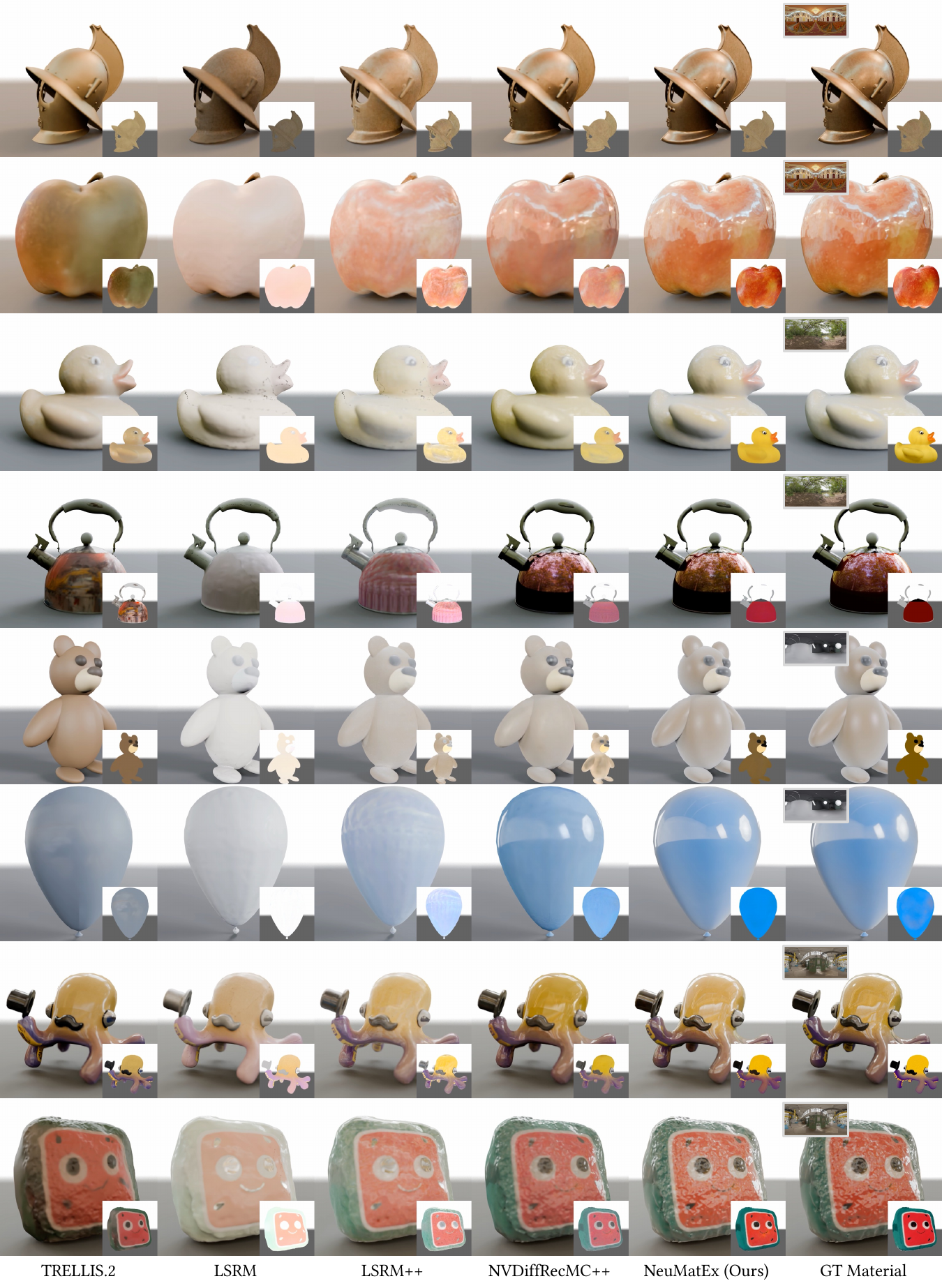}
\caption{\textbf{PBR vs. Neural Material (128~spp, relit).}
Insets: extracted diffuse base color (all columns, bottom), probes for relighting (last column, top).
}
\label{fig:supp_comparison}
\end{figure*}

\clearpage

\subsection{More Real-world Neural Material Extraction Results}

\begin{figure*}[ht]
\centering
\includegraphics[width=\linewidth]{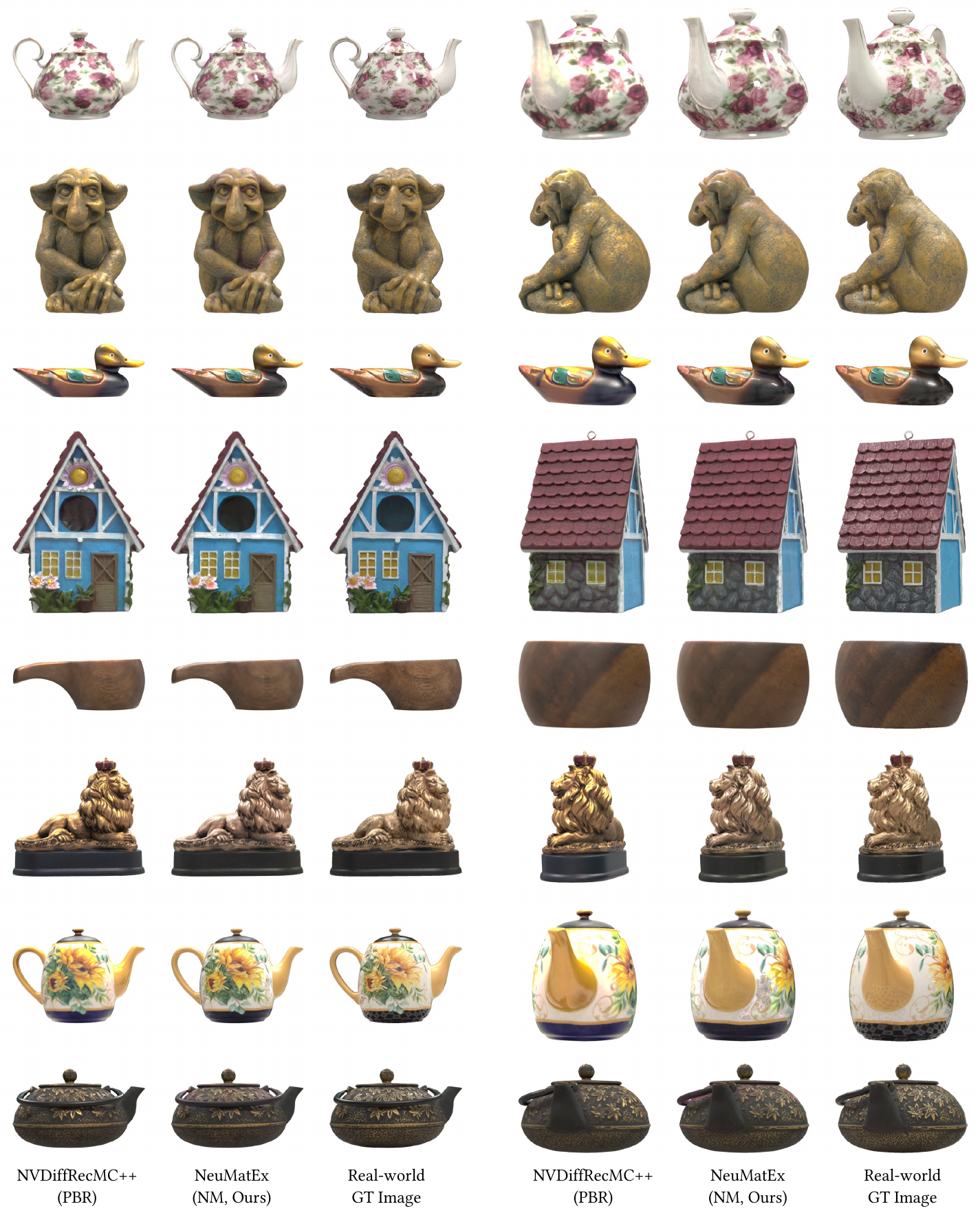}
\caption{\textbf{Comparison for real-world neural material extraction.}
Additional results on the DTC~\citep{dong2025dtc} dataset. Our method extracts more expressive neural materials than the PBR-based optimization method NVDiffRecMC++, capturing complex specular reflections.
}
\label{fig:realworld:supp:comparison}
\end{figure*}

\clearpage

We quantitatively compare the relighting performance of NVDiffRecMC++ and \ours, which indicates the quality of material decomposition.
Note that the DTC benchmark is not designed with neural material reconstruction in mind: its ground-truth mesh assets are annotated with PBR materials, so the relit reference renderings cannot exceed the expressiveness of the PBR model.
This setting inherently favors PBR-based methods, and we therefore expect only a small gap between the two approaches.
Nevertheless, our extracted materials better capture specular effects and outperform NVDiffRecMC++ on all relighting metrics in \Tref{tab:dtc_relighting}.

\begin{table}[h]
\centering
\small
\setlength{\tabcolsep}{8pt}
\caption{
  Relighting performance on the DTC benchmark.
  Results are averaged over four HDR environments and eight assets.
  Best results are shown in bold.
}
\label{tab:dtc_relighting}
    \begin{tabular}{lccc}
      \toprule
      \multirow{2}{*}{Method}
      & \multicolumn{3}{c}{Relighting} \\
      \cmidrule(lr){2-4}
      & PSNR $\uparrow$
      & LPIPS $\downarrow$
      & SSIM $\uparrow$ \\
      \midrule
      \textsc{NVDiffRecMC++}
      & 23.70
      & 0.039
      & 0.951 \\
      \textsc{\ours} (Ours, Full)
      & \textbf{24.03}
      & \textbf{0.037}
      & \textbf{0.952} \\
      \bottomrule
    \end{tabular}
\end{table}

\begin{figure*}[ht]
\centering
\includegraphics[width=\linewidth]{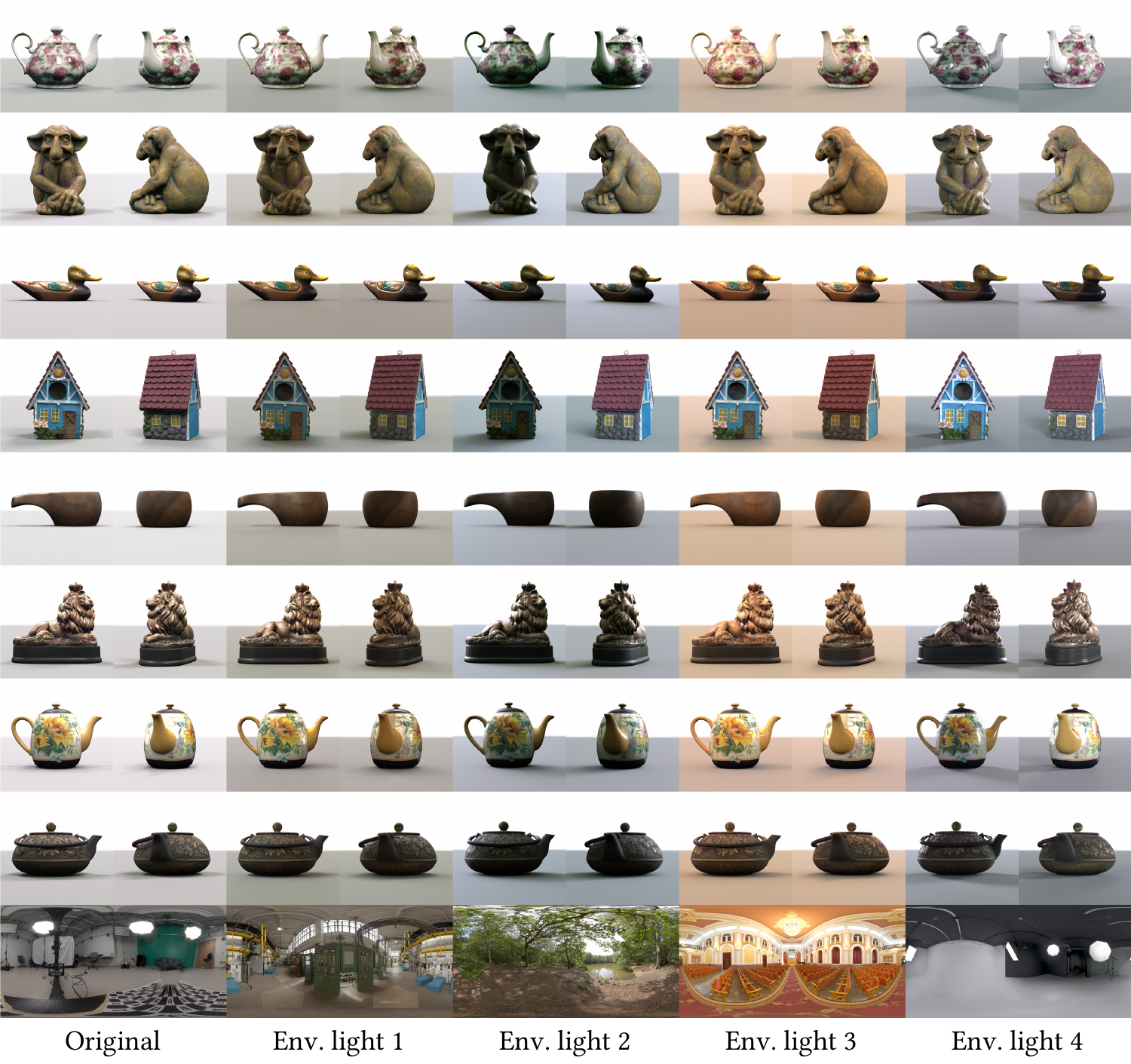}
\caption{\textbf{Relighting real-world neural materials.}
Relighting results showing two views of multiple objects from the DTC~\citep{dong2025dtc} dataset.
Our extracted neural materials generalize to novel lighting conditions producing complex specular effects. 
}
\label{fig:realworld:supp:relighting}
\end{figure*}

\end{document}